\documentclass[sn-mathphys,iicol]{sn-jnl}

\usepackage{graphicx}%
\usepackage{multirow}%
\usepackage{amsmath,amssymb,amsfonts}%
\usepackage{amsthm}%
\usepackage{mathrsfs}%
\usepackage[title]{appendix}%
\usepackage{xcolor}%
\usepackage{textcomp}%
\usepackage{manyfoot}%
\usepackage{booktabs}%
\usepackage{algorithm}%
\usepackage{algorithmicx}%
\usepackage{algpseudocode}%
\usepackage{listings}%

\usepackage{xcolor} %
\usepackage{graphicx} %
\usepackage{booktabs} %
\usepackage{multirow} %
\usepackage{algorithm} %
\usepackage{algpseudocode} %
\usepackage{inconsolata} %

\usepackage{anyfontsize} %

\definecolor{light-gray}{HTML}{b7b7b7} %
\definecolor{unseen}{HTML}{6aa84f} %
\definecolor{highlight}{HTML}{cfe2f3} %

\renewcommand\fbox{\fcolorbox{light-gray}{white}}

\newcommand{\llmcompletion}[1]{\colorbox{highlight}{#1}}
\newcommand{\codeblock}[1]{\vspace{6pt}\noindent\fbox{\parbox{0.99\linewidth}{\small{\raggedright\texttt{#1}}}}\vspace{6pt}}

\newcommand{\ie}{\textit{i.e.}, }
\newcommand{\eg}{\textit{e.g.}, }
\newcommand{\mysubsection}[1]{\vspace{6pt}\noindent{\textbf{#1}}}

\begin{document}

\title[TidyBot: Personalized Robot Assistance with Large Language Models]{TidyBot: Personalized Robot Assistance with \mbox{Large Language Models}}

\author*[1]{\fnm{Jimmy} \sur{Wu}}\email{jw60@cs.princeton.edu}
\author[2]{\fnm{Rika} \sur{Antonova}}\email{rika.antonova@stanford.edu}
\author[3]{\fnm{Adam} \sur{Kan}}\email{adakan@nuevaschool.org}
\author[2]{\fnm{Marion} \sur{Lepert}}\email{lepertm@stanford.edu}
\author[4]{\fnm{Andy} \sur{Zeng}}\email{andyzeng@google.com}
\author[5]{\fnm{Shuran} \sur{Song}}\email{shurans@cs.columbia.edu}
\author[2]{\fnm{Jeannette} \sur{Bohg}}\email{bohg@stanford.edu}
\author[1]{\fnm{Szymon} \sur{Rusinkiewicz}}\email{smr@princeton.edu}
\author[1,4]{\fnm{Thomas} \sur{Funkhouser}}\email{funk@cs.princeton.edu}
\affil*[1]{\orgname{Princeton University}, \orgaddress{\city{Princeton}, \state{NJ}, \country{USA}}}
\affil[2]{\orgname{Stanford University}, \orgaddress{\city{Stanford}, \state{CA}, \country{USA}}}
\affil[3]{\orgname{The Nueva School}, \orgaddress{\city{San Mateo}, \state{CA}, \country{USA}}}
\affil[4]{\orgname{Google}, \orgaddress{\city{Mountain View}, \state{CA}, \country{USA}}}
\affil[5]{\orgname{Columbia University}, \orgaddress{\city{New York}, \state{NY}, \country{USA}}}

\abstract{For a robot to personalize physical assistance effectively, it must learn user preferences that can be generally reapplied to future scenarios. In this work, we investigate personalization of household cleanup with robots that can tidy up rooms by picking up objects and putting them away. A key challenge is determining the proper place to put each object, as people's preferences can vary greatly depending on personal taste or cultural background. For instance, one person may prefer storing shirts in the drawer, while another may prefer them on the shelf. We aim to build systems that can learn such preferences from just a handful of examples via prior interactions with a particular person. We show that robots can combine language-based planning and perception with the few-shot summarization capabilities of large language models (LLMs) to infer generalized user preferences that are broadly applicable to future interactions. This approach enables fast adaptation and achieves 91.2\% accuracy on unseen objects in our benchmark dataset. We also demonstrate our approach on a real-world mobile manipulator called TidyBot, which successfully puts away 85.0\% of objects in real-world test scenarios.}


\maketitle

\section{Introduction}

Building a robot that provides personalized assistance for physical household tasks is a long-standing goal of robotics research.
In this paper, we investigate the task of tidying up a room: moving every object on the floor to its ``proper place.''
One of the challenges in performing this task is determining the correct receptacle (``proper place'') for every object.
This is difficult because where objects should go is highly personal, and depends on cultural norms and individual preferences.
One person may want to put shirts in a dresser drawer, another may want them on shelves, and a third may want them hanging in a closet.
There is no ``one size fits all'' solution.

Classical approaches to the household cleanup task ask a person to specify a target location for every object~\citep{rasch2019tidy, yan2021quantifiable}, which is tedious and impractical in an autonomous setting.
Other works learn generic (non-personalized) rules about where objects typically go inside a house by averaging over many users~\citep{taniguchi2021autonomous,kant2022housekeep,sarch2022tidee}.
Works that focus on personalization aim to extrapolate from a few user examples given similar choices made by other users, using methods such as collaborative filtering~\citep{abdo2015robot}, spatial relationships~\citep{kang2018automated}, or learned latent preference vectors~\citep{kapelyukh2022my}.
However, all of these approaches require collecting large datasets with user preferences or generating datasets from manually constructed, simulated scenarios.
Such datasets can be expensive to acquire and may not generalize well if they are too small.

Our approach is to utilize the summarization capabilities of large language models (LLMs) to provide generalization from a small number of example preferences.
We ask a person to provide a few example object placements using textual input (\eg yellow shirts go in the drawer, dark purple shirts go in the closet, white socks go in the drawer), and then we ask the LLM to summarize these examples (\eg light-colored clothes go in the drawer and dark-colored clothes go in the closet) to arrive at  generalized preferences for this particular person.

The underlying insight is that the summarization capabilities of LLMs are a good match for the generalization requirements of personalized robotics.
LLMs demonstrate astonishing abilities to perform generalization through summarization, drawing upon complex object properties and relationships learned from massive text datasets.
By using the summarization provided by LLMs for generalization in robotics, we hope to produce generalized rules from a small number of examples, in a form that is human interpretable (text) and is expressed in nouns that can be grounded in images using open-vocabulary image classifiers.
Using an off-the-shelf LLM also avoids expensive collection of user preference data and model training.

\begin{figure}
\centering
\includegraphics[width=\columnwidth]{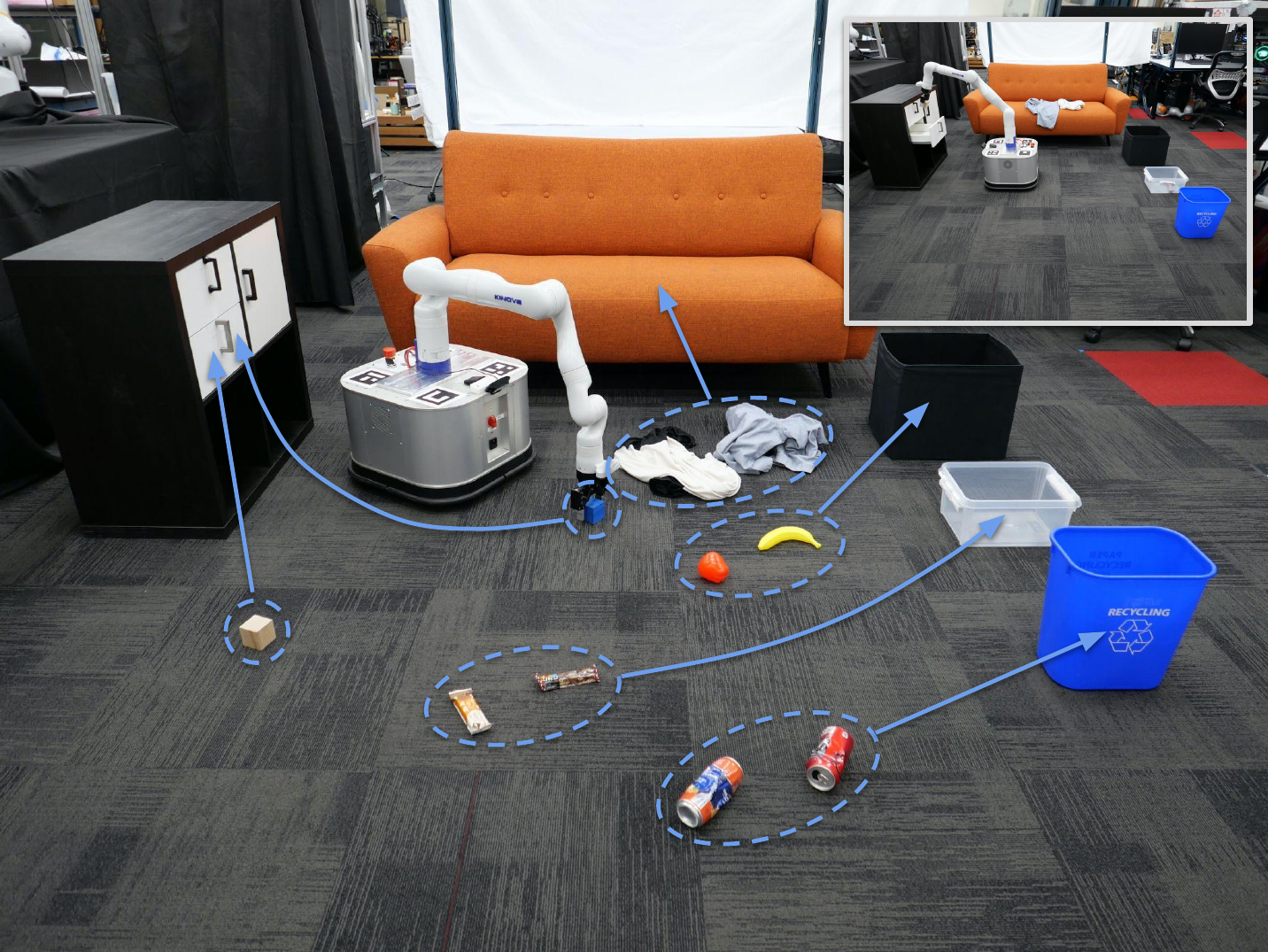}
\caption{We study the task of household cleanup, where each object on the floor must be picked up and put away while following user preferences.}
\label{fig:teaser}
\end{figure}

We investigate the proposed approach in a real-world robotic mobile manipulation system for household cleanup, which we call TidyBot (Fig.~\ref{fig:teaser}).
Before the robot begins cleanup, we ask the user to provide a handful of example placements for specific objects, which are passed to an LLM to be summarized into a generalized set of rules (personalized to that user) mapping object categories to receptacles.
The nouns of these generalized rules are provided to an open-vocabulary image classifier in order to identify objects on the floor and determine target receptacles for them using the rules.
The robot will then carry out the cleanup task by repeatedly picking up objects, identifying them, and moving them to their target receptacles.

We evaluate our approach quantitatively on both a text-based benchmark dataset and our real-world robotic system.
On the benchmark, we find that our approach generalizes well, achieving an accuracy of 91.2\% on unseen objects across all scenarios in the benchmark.
In our real-world test scenarios, we find that TidyBot correctly puts away 85.0\% of objects.
We also show that our approach can be easily extended to infer generalized rules for manipulation primitive selection (\eg pick and place vs. pick and toss) in addition to inferring object placements.

Our contributions are:
(i) the idea that text summarization with LLMs provides a means for generalization in robotics,
(ii) a publicly released benchmark dataset for evaluating generalization of receptacle selection preferences, and
(iii) implementation and evaluation of our approach on a real-world mobile manipulation system.

This journal paper is an extended version of a previously published conference paper~\citep{wu2023tidybot}. The new material in this journal version includes:

\begin{enumerate}
\item A user study that evaluates whether humans prefer the preferences learned by our approach, and whether human responses align with our benchmark's ground truth
\item Quantitative analysis of the perception component of the real-world system, including comparisons of different visual language models
\item Additional statistics of our benchmark showing representation of different sorting criteria in the dataset, along with a breakdown of baseline results according to these criteria
\item A summary of the limitations of our system
\end{enumerate}

Please see our project page at \url{https://tidybot.cs.princeton.edu} for additional supplementary material, benchmark dataset and code, and qualitative videos of our real-world system TidyBot in action.

\section{Related Work}

\mysubsection{Household cleanup.}
Many recent works in Embodied AI have
proposed benchmarks or methods for completing household tasks in simulated indoor environments~\citep{kolve2017ai2,puig2018virtualhome,shridhar2020alfred,shridhar2021alfworld,szot2021habitat,li2022igibson,srivastava2022behavior,li2022behavior}.
For household cleanup in particular, the object rearrangement task~\citep{puig2018virtualhome,batra2020rearrangement,szot2021habitat,ehsani2021manipulathor,weihs2021visual,gan2022threedworld} requires an embodied agent to pick up and move objects so as to bring the environment into a specified state.
Household cleanup has also been studied in robotics works, in which instructions for object rearrangement are specified via pointing gestures~\citep{rasch2019tidy} or target layouts~\citep{yan2021quantifiable}.
The drawback of these setups is that a target location must be manually specified for every object to be manipulated, which can require significant human effort.
Prior works have addressed this challenge by automatically inferring object placements based on human preferences for where objects typically go inside a house~\citep{taniguchi2021autonomous,kant2022housekeep,sarch2022tidee}, eliminating the need to specify where every individual object goes.
However, these works predict human preferences that are generic rather than personalized.
To handle the variability in preferences across different users, other works have used collaborative filtering~\citep{abdo2015robot}, spatial relationships~\citep{kang2018automated}, or learned latent preference vectors~\citep{kapelyukh2022my} to predict object placements that are based on personalized user preferences.
These methods require the collection of large crowd-sourced datasets for human preferences, which can be expensive.
By contrast, our approach uses off-the-shelf LLMs with no additional training or data collection.
We are able to directly leverage the commonsense knowledge and summarization abilities of LLMs to build generalizable personalized preferences for each user.

\mysubsection{Object sorting.}
Object sorting has been studied in robotics using approaches such as clustering~\citep{gupta2012using}, active learning~\citep{kujala2016classifying,herde2018active}, metric learning~\citep{zeng2022robotic}, or heuristic search~\citep{huang2019large,song2020multi,pan2021decision}.
These setups carry out pre-specified sorting rules using physical properties such as color~\citep{szabo2012automated,gupta2012using,kujala2016classifying,herde2018active,huang2019large,dewi2020fruit,song2020multi,pan2021decision}, shape~\citep{herde2018active}, size~\citep{gupta2012using,herde2018active,dewi2020fruit}, or material~\citep{lukka2014zenrobotics}.
Notably, they are not able to sort based on semantics or commonsense knowledge, nor are they able to automatically infer sorting rules.
More recently, ~\cite{hoeg2022more} studied whether classification of objects into general high-level categories can be improved by using an LLM to take in an object detector's prediction and output a general category for the object.
In our work, we similarly tap into the commonsense knowledge of LLMs to reason about object sorting.
However, whereas their setup uses pre-specified sorting rules based on a fixed set of categories, ours is able to infer generalizable sorting rules automatically.

\mysubsection{LLMs for robotics.}
Large language models (LLMs) have been shown to exhibit remarkable commonsense reasoning abilities~\citep{brown2020language,nye2021show,rytting2021leveraging,wei2022chain,wei2022emergent,kojima2022large,madaan2022language}.
As a result, there has been increasing interest in harnessing the capabilities of LLMs to build more commonsense knowledge into robotic systems.
Many recent works study how LLM-generated high-level robotic plans (typically produced using the few-shot learning paradigm~\citep{brown2020language}) can be grounded in the state of the environment.
This can be done with value functions~\citep{brohan2022can,lin2023text2motion}, semantic translation into admissible actions~\citep{huang2022language}, scene description as context~\citep{zeng2022socratic,mees2022grounding,chen2022open,singh2022progprompt}, feedback~\citep{huang2022inner,yao2022react}, or re-prompting~\citep{raman2022planning}.
However, these works assume a setup in which the LLM is expected to output a single generic plan. This is not a good fit for personalized household cleanup, because a ``one size fits all'' plan would not address the wide variability in user preferences.
Instead, our system generates personalized plans that are tailored to the preferences of a particular user.
Other works in robotics have used LLMs for PDDL planning~\citep{silver2022pddl}, code generation for robotic control policies~\citep{liang2022code}, parsing navigation instructions into textual landmarks~\citep{shah2022lm}, room classification~\citep{chen2022leveraging}, and tool manipulation~\citep{ren2022leveraging}.
These works all use LLMs as a means of integrating commonsense knowledge into robotic systems, which is also true in our case.
However, unlike these works, we additionally show that the summarization ability of LLMs enables generalization in robotics.

\section{Method}

We use the summarization capabilities of an off-the-shelf LLM to generalize user preferences from a small number of examples.
Below, we describe how we use the LLM to infer personalized rules for both receptacle selection and manipulation primitive selection, and also how we deploy the approach on a real-world mobile manipulation system for household cleanup.

\subsection{Personalized receptacle selection}

Our system first receives a few examples of object placements reflecting the personal preferences of a user.
For instance, the user may specify that yellow shirts and white socks go in the drawer, while dark purple shirts and black shirts go in the closet.
We provide these examples to an LLM, which then infers personalized rules on where objects belong.
Specifically, the LLM (i) summarizes the examples into general rules, and then (ii) uses the summary to determine where to place new objects.

Following recent work~\citep{zeng2022socratic,singh2022progprompt}, we convert the user examples into LLM prompts that are structured as Pythonic code.
This prompt form is advantageous because LLMs are trained on large amounts of code, and it also provides a structured output that is easy to parse.
To represent the user examples, the prompt first contains a list of objects present in the scene and a list of potential receptacles (see Appendix~\ref{sec:appendix-full-prompts} for full prompt with in-context examples).
This is followed by a series of pick and place commands reflecting where the objects should be placed according to the user.
Then, we ask the LLM to complete the last line, which is a code comment summarizing what the preceding code block does.
Here is an example LLM completion where the output from the LLM is \colorbox{highlight}{highlighted}:

\codeblock{objects = ["yellow shirt", "dark purple shirt", "white socks", "black shirt"]\\
receptacles = ["drawer", "closet"]\\
pick\_and\_place("yellow shirt", "drawer")\\
pick\_and\_place("dark purple shirt", "closet")\\
pick\_and\_place("white socks", "drawer")\\
pick\_and\_place("black shirt", "closet")\\
\# Summary:\llmcompletion{ Put light-colored clothes in the }\\
\llmcompletion{drawer and dark-colored clothes in the closet.}}

In this example, the LLM summarized the provided object placements and inferred that light-colored clothes go in the drawer while dark-colored clothes go in the closet.
These examples lead to a generalized rule for where objects belong, personalized to this particular user.

Next, the summary is used by the LLM to generate placements for novel, unseen objects.
The prompt consists of the summary from the LLM summarization step (in the form of a code comment), a list of the \textcolor{unseen}{unseen objects}, a list of receptacles, and a partial pick and place command for the first object.
We then ask the LLM to provide a placement for each object by completing the prompt:

\codeblock{\# Summary: Put light-colored clothes in the drawer and dark-colored clothes in the closet.\\
objects = [\textcolor{unseen}{"black socks", "white shirt", "navy socks", "beige shirt"}]\\
receptacles = ["drawer", "closet"]\\
pick\_and\_place("black socks",\llmcompletion{ "closet")}\\
\llmcompletion{pick\_and\_place("white shirt", "drawer")}\\
\llmcompletion{pick\_and\_place("navy socks", "closet")}\\
\llmcompletion{pick\_and\_place("beige shirt", "drawer")}}

The output pick and place commands can then be parsed to determine where each unseen object should be placed.

\subsection{Personalized primitive selection}

Similar to the way we infer generalized rules for receptacle selection, we can also infer generalized rules for how to manipulate objects, again leveraging the summarization capabilities of LLMs.
First, we provide a few examples of objects along with their user-preferred manipulation primitive to the LLM, and ask it to summarize.
Here is an example completion where the output from the LLM is \colorbox{highlight}{highlighted}:

\codeblock{objects = ["yellow shirt", "dark purple shirt", "white socks", "black shirt"]\\
pick\_and\_place("yellow shirt")\\
pick\_and\_place("dark purple shirt")\\
pick\_and\_toss("white socks")\\
pick\_and\_place("black shirt")\\
\# Summary:\llmcompletion{ Pick and place shirts, pick and }\\
\llmcompletion{toss socks.}}

The summary can then be used as a generalized rule to predict the appropriate primitive to use for \textcolor{unseen}{unseen objects}:

\codeblock{\# Summary: Pick and place shirts, pick and toss socks.\\
objects = [\textcolor{unseen}{"black socks", "white shirt", "navy socks", "beige shirt"}]\\
\llmcompletion{pick\_and\_toss("black socks")}\\
\llmcompletion{pick\_and\_place("white shirt")}\\
\llmcompletion{pick\_and\_toss("navy socks")}\\
\llmcompletion{pick\_and\_place("beige shirt")}}

\subsection{Real-world robotic system}
\label{sec:system}

Given generalized rules from LLM summarization, we can now implement these rules on a robot tasked with tidying up a household environment.
To do so, we use a perception system to localize and recognize objects in the environment, and a predetermined set of manipulation primitives to move objects into receptacles. For our setup, we use \texttt{pick\_and\_place} and \texttt{pick\_and\_toss} as our primitives, as they are well-suited for household cleanup. However, other sets of primitives could also be used.

\begin{figure*}
\includegraphics[width=\textwidth]{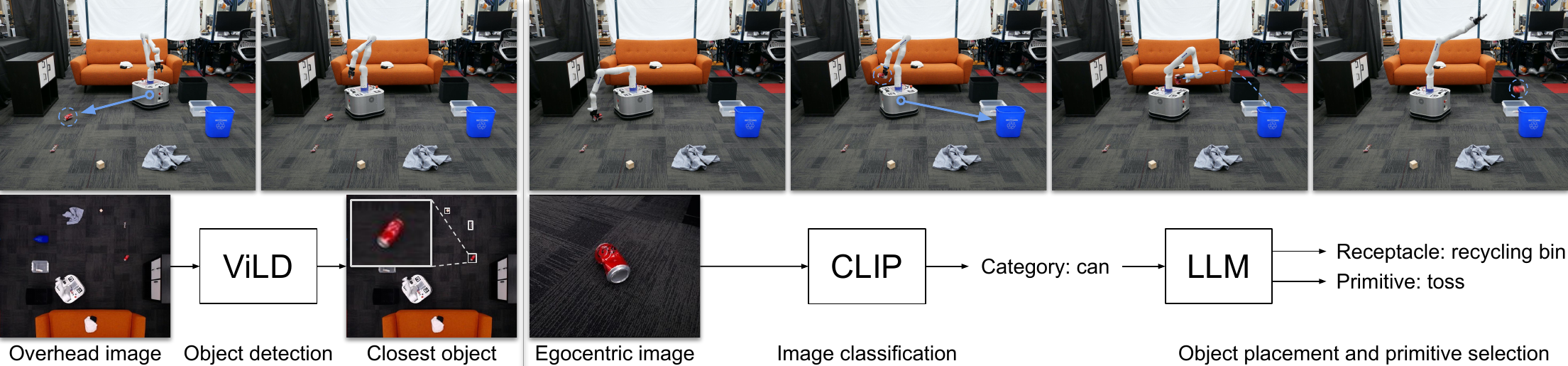}
\caption{\textbf{System overview.} Once the user's preferences have been summarized with an LLM, TidyBot will localize the closest object on the floor, move to get a close-up view with its egocentric camera, predict the object's category using CLIP, use the LLM-summarized rules to select a receptacle and manipulation primitive, and then execute the primitive to put the object into the selected receptacle, repeating this entire process until no more objects can be found on the floor.}
\label{fig:overview}
\end{figure*}

For each new user, the system will receive a set of example preferences and run the previously described LLM pipeline to get personalized rules for the user.
The rules contain a set of generalized object categories produced by summarization (\eg light-colored clothes, dark-colored clothes), each of which is matched to a preferred receptacle and manipulation primitive for that category.
The robot will tidy up the environment by iteratively performing the following steps until no more objects remain on the floor: (1) localize the nearest object, (2) classify the object into a generalized category, (3) determine the appropriate receptacle and manipulation primitive for the object using generalized rules produced by the LLM, and (4) use the manipulation primitive to put the object into the receptacle.
Fig.~\ref{fig:overview} provides a conceptual illustration of this procedure, and Algorithm~\ref{alg:system-pipeline} outlines these steps in pseudocode.

\begin{algorithm}
\caption{System pipeline}\label{alg:system-pipeline}
\begin{algorithmic}
\State \textbf{Input}: $E_\mathrm{receptacle}$ = $\{(o_1, r_1), (o_2, r_2), \ldots\}$
\State \textbf{Input}: $E_\mathrm{primitive}$ = $\{(o_1, p_1), (o_2, p_2), \ldots\}$
\State $S_\mathrm{receptacle}$ = \textit{LLM}.Summarize($E_\mathrm{receptacle}$)
\State $S_\mathrm{primitive}$ = \textit{LLM}.Summarize($E_\mathrm{primitive}$)
\State $C$ = \textit{LLM}.GetCategories($S_\mathrm{receptacle}$)
\State \textit{robot}.Initialize()
\While{True}
\State $I_\mathrm{top}$ = GetOverheadImage()
\State $o$ = \textit{ViLD}.GetClosestObject($I_\mathrm{top}$)
\State \textit{robot}.MoveTo($o$)
\State $I_\mathrm{ego}$ = \textit{robot}.GetEgocentricImage()
\State $c$ = \textit{CLIP}.GetCategory($I_\mathrm{ego}$, $C$)
\State $r$ = \textit{LLM}.GetReceptacle($S_\mathrm{receptacle}$, $c$)
\State $p$ = \textit{LLM}.GetPrimitive($S_\mathrm{primitive}$, $c$)
\State \textit{robot}.PickUp($o$)
\State \textit{robot}.MoveTo($r$)
\State \textit{robot}.ExecutePrimitive($p$)
\EndWhile
\end{algorithmic}
\end{algorithm}

One important aspect of our approach is that the LLM summarization automatically provides candidate categories to the perception system.
Nouns (or noun phrases) are extracted from the summarization text as categories, and used as the target label set for CLIP~\citep{radford2021learning}, the open-vocabulary image classification model we use.
For example, the following LLM prompt will extract the two general categories in the summary text (light-colored clothing and dark-colored clothing):

\codeblock{\# Summary: Put light-colored clothes in the drawer and dark-colored clothes in the closet.\\
objects = ["\llmcompletion{light-colored clothing", }\\
\llmcompletion{"dark-colored clothing"]}}

This combination of summarization and open-vocabulary classification is critical to the autonomy of the system, as it enables the object classifier to work with a small set of generalized object categories.
The approach is (i) robust as there are only a small number of categories to differentiate between, and (ii) flexible because it supports arbitrary sets of object categories for different users.
In contrast, without LLM summarization, the object classifier would have to be able to recognize all possible fine-grained object classes, which is much more difficult.
Alternatively, the user would have to manually specify the list of objects present in each target scene, which would be impractical for an autonomous system.

\section{Experiments}

We investigate the performance of our proposed approach with two types of evaluation.
For the first type of evaluation, we design a benchmark for generalization of receptacle selection using text-based examples, which enables direct comparison to alternative approaches and ablation studies, with quantitative metrics.
For the second type of evaluation, we deploy our approach in a real-world mobile manipulation system for tidying up a room based on user preferences.
Unless otherwise specified, the LLM we use is \texttt{text-davinci-003}, a variant of GPT-3~\citep{brown2020language}.
All LLM experiments were run with temperature 0.

\subsection{Benchmark dataset}
\label{sec:benchmark}

In order to evaluate the proposed approach and to quantitatively compare it to alternatives, we created a benchmark dataset of object placements.
The benchmark is comprised of 96 scenarios, each of which has a set of objects, a set of receptacles, a set of example ``seen'' object placements (preferences), and a set of ``unseen'' evaluation placements, all specified as text.
The task is to predict the placements in the ``unseen'' set given the examples in the ``seen'' set.

The benchmark scenarios are defined in 4 room types (living room, bedroom, kitchen, pantry room), with 24 scenarios per room type.
Each scenario contains 2--5 receptacles (potential places to put objects,  such as shelves, cabinets, etc.), 4--10 ``seen'' example object placements provided as input to the task, and an equal number of ``unseen'' object placements (distinct from the seen examples) provided for evaluation.
There are 2 seen and 2 unseen object placements per receptacle.
In total, there are 672 seen and 672 unseen object placements, which cumulatively reference 87 unique receptacles and 1,076 unique objects.

Success on this benchmark is measured by the object placement accuracy: the number of objects placed in the correct receptacle divided by the total number of objects.  We evaluate accuracy separately for seen and unseen objects, to tease apart memorization versus generalization.
For each, we compute the accuracy per scenario, and then average the results across all scenarios to produce the numbers shown in the tables.

Since different people may sort items in the home in many different ways, our benchmark contains a diversity of preferences with several kinds of sorting criteria represented in the dataset:

\begin{itemize}
\item \textbf{Category:} Sort objects based on general categories (\eg put clothes here and toys there)
\item \textbf{Attribute:} Sort objects based on object attributes (\eg put plastic items here and metal items there)
\item \textbf{Function:} Sort objects based on function (\eg put winter clothes here and summer clothes there)
\item \textbf{Subcategory:} Sort objects such that a specific (subordinate) subcategory is separated from the general (superordinate) category (\eg put shirts on the sofa and other clothes in the closet)
\item \textbf{Multiple categories:} Sort objects from multiple categories into one receptacle (\eg put both books and toys on the shelf)
\end{itemize}

We show in Tab.~\ref{tab:benchmark-criteria} the representation of different sorting criteria in our benchmark dataset, indicated by the fraction of the 96 scenarios to which each criteria applies. Note that multiple sorting criteria may apply to a single scenario.

\begin{table}
\caption{Representation of sorting criteria in benchmark}
\centering
\small
\setlength\tabcolsep{0.22em}
\begin{tabular}{ccccc}
\toprule
Category & Attribute & Function & Subcategory & Multiple \\
\midrule
86/96 & 27/96 & 24/96 & 31/96 & 17/96 \\
\bottomrule
\end{tabular}
\label{tab:benchmark-criteria}
\end{table}

\subsection{Baseline comparisons}
\label{sec:baseline-comparisons}

In our first set of experiments, we use the benchmark to provide quantitative evaluation of our approach compared to several alternatives.
The results are in Tab.~\ref{tab:baseline-comparison}.
We also show in Tab.~\ref{tab:baseline-comparison-criteria} the same results but broken down by the sorting criteria described in Sec.~\ref{sec:benchmark}.
Since the main challenge is to generalize from objects in the examples (seen) to those in the evaluation set (unseen), we consider a variety of baseline generalization approaches and report placement accuracy metrics only for unseen objects.

\begin{table}
\caption{Comparisons to baselines}
\centering
\small
\begin{tabular}{l|c}
\toprule
Method & Accuracy (unseen) \\
\midrule
Examples only & 78.5\% \\
WordNet taxonomy & 67.5\% \\
RoBERTa embeddings & 77.8\% \\
CLIP embeddings & 83.7\% \\
Summarization (ours) & \textbf{91.2\%} \\
\bottomrule
\end{tabular}
\label{tab:baseline-comparison}
\end{table}

\begin{table*}
\caption{Comparisons to baselines by sorting criteria}
\centering
\small
\setlength\tabcolsep{1.5em}
\begin{tabular}{l|ccccc}
\toprule
Method & Category & Attribute & Function & Subcategory & Multiple \\
\midrule
Examples only        & 80.1\% & 72.7\% & 75.7\% & 77.0\% & 81.5\% \\
WordNet taxonomy     & 69.1\% & 59.8\% & 61.4\% & 71.3\% & 74.1\% \\
RoBERTa embeddings   & 78.6\% & 75.5\% & 71.8\% & 71.7\% & 87.5\% \\
CLIP embeddings      & 84.6\% & 79.8\% & 85.5\% & 84.7\% & 87.9\% \\
Summarization (ours) & \textbf{91.0\%} & \textbf{85.6\%} & \textbf{93.9\%} & \textbf{90.1\%} & \textbf{93.5\%} \\
\bottomrule
\end{tabular}
\label{tab:baseline-comparison-criteria}
\end{table*}

The following paragraphs describe each baseline and provide a discussion of how the performance compares to that of our proposed approach.

\mysubsection{Examples only.}
The first baseline provides a direct comparison to a system like ours if it did not use summarization.
The LLM is given a list of objects, receptacles, and example placement preferences, along with a list of unseen objects for a new scene. Then, the LLM is asked to directly infer the proper placements (\colorbox{highlight}{highlighted} text) for \textcolor{unseen}{unseen objects} in the new scene, without summarization as an intermediate step:

\codeblock{objects = ["yellow shirt", "dark purple shirt", "white socks", "black shirt"]\\
receptacles = ["drawer", "closet"]\\
pick\_and\_place("yellow shirt", "drawer")\\
pick\_and\_place("dark purple shirt", "closet")\\
pick\_and\_place("white socks", "drawer")\\
pick\_and\_place("black shirt", "closet")\\
\phantom{}\\
objects = [\textcolor{unseen}{"black socks", "white shirt", "navy socks", "beige shirt"}]\\
receptacles = ["drawer", "closet"]\\
pick\_and\_place("black socks",\llmcompletion{ "drawer")}\\
\llmcompletion{pick\_and\_place("white shirt", "closet")}\\
\llmcompletion{pick\_and\_place("navy socks", "drawer")}\\
\llmcompletion{pick\_and\_place("beige shirt", "closet")}}

The prediction accuracy of this method for unseen objects (78.5\%) is significantly worse than that of our method (91.2\%). Since the main difference between this method versus ours is that our method leverages summarization, this result presents strong evidence for our main hypothesis --- \ie summarization is useful for generalization.
This finding is also consistent with recent work showing that LLMs perform better when they are asked to output intermediate steps of reasoning before the final answer~\citep{nye2021show,wei2022chain}.
When looking at the predictions, we find that this baseline approach generally predicts object placements that are sensible but may not be consistent with the user's preferences.

\mysubsection{WordNet taxonomy.}
This baseline uses a hand-crafted lexical ontology called WordNet~\citep{miller1995wordnet} to generalize placements from seen to unseen objects.
For each unseen object, we place it in the same receptacle as the most similar seen object, where similarity is measured using the shortest path between two objects in the taxonomy.
Since WordNet is a hand-crafted taxonomy, it does not contain all possible object names.
For the 694 objects in our benchmark that are missing from WordNet, we manually mapped each of them to the closest WordNet object name.
Even with the manual mapping, the performance of this WordNet baseline for unseen objects (67.5\%) is far worse than that of our method (91.2\%).
This shows that LLM summarization provides better generalization than using the hierarchy provided by a hand-crafted ontology.
When looking at the breakdown in Tab.~\ref{tab:baseline-comparison-criteria}, we see that this baseline performs worse on the two criteria that are not related to object categorization (attribute and function).
We hypothesize that WordNet is not able to generalize well along these dimensions because it was constructed mainly based on semantic relationships between categories.

\mysubsection{Text embedding.}
This baseline uses pretrained text embeddings to assist with generalization.
For each unseen object, we place it in the receptacle provided for the most similar seen object, where similarity is defined by cosine similarity between encoded object names in the RoBERTa~\citep{liu2019roberta} or CLIP~\citep{radford2021learning} embedding space.
For RoBERTa, we use the pretrained Sentence-BERT~\citep{reimers2019sentence} model from the SentenceTransformers library.
Specifically, we use the \texttt{all-distilroberta-v1} variant which is a distilled~\citep{sanh2019distilbert} version of the RoBERTa~\citep{liu2019roberta} model that is fine-tuned on a dataset of 1 billion sentence pairs.
For CLIP, we use the pretrained model provided by OpenAI.
In either case, the generalization performance for predicting placements of unseen objects does not reach the performance of our proposed summarization approach (77.8\% for RoBERTa and 83.7\% for CLIP, versus 91.2\% for ours).
Although text embeddings trained on large datasets encode many types of object similarities, particularly for related object categories, they may not encode the object attributes relevant to the preferences of a particular user (\eg light objects go here, heavy object go there).
In contrast, our summarization approach is able to correctly encode a larger variety of user preferences.

\subsection{Ablation studies}

In the second set of experiments, we use the benchmark to evaluate the performance of several variants to our method.
The goal of these experiments is to compare its performance to alternatives with far less information (using only common sense, without preferences) or far more information (using human-generated summarizations).
We also study the impact of using different LLMs.
The benchmark metrics for both seen and unseen objects are provided in Tabs.~\ref{tab:ablations} and \ref{tab:llm-comparison}.

\begin{table}
\caption{Ablation studies}
\centering
\small
\begin{tabular}{l|cc}
\toprule
Method & Seen & Unseen \\
\midrule
Commonsense & 45.0\% & 45.6\% \\
Summarization & 91.8\% & 91.2\% \\
Human summary & 97.1\% & 97.5\% \\
\bottomrule
\end{tabular}
\label{tab:ablations}
\end{table}

\mysubsection{Commonsense.}
Our first ablation study measures how well an LLM can perform
the benchmark tasks using only commonsense reasoning --- \ie without using the preferences at all.
For each benchmark scene, we give the LLM the list of objects and list of receptacles, and then ask it to generate object placements (\colorbox{highlight}{highlighted} text) without using the provided user preferences:

\codeblock{\# Put objects into their appropriate receptacles.\\
objects = ["black socks", "white shirt", "navy socks", "beige shirt"]\\
receptacles = ["drawer", "closet"]\\
pick\_and\_place("black socks",\llmcompletion{ "drawer")}\\
\llmcompletion{pick\_and\_place("white shirt", "closet")}\\
\llmcompletion{pick\_and\_place("navy socks", "drawer")}\\
\llmcompletion{pick\_and\_place("beige shirt", "closet")}}

This baseline performs poorly, even for seen objects (45.0\%), due to the high variability of object placement preferences in the benchmark.
The predicted object placements are sensible but are not reflective of the particular user's preferences.
In contrast, our method can learn preferences from examples via summarization and performs much better for both seen and unseen objects
(91.8\% and 91.2\%).

\mysubsection{Human summary.}
This ablation studies how the summaries provided by the LLM compare to summaries crafted manually by a human.
For each benchmark scenario, a human-written summary was used by the LLM (in place of the LLM-produced summary) to predict object placements for the test objects.
The results achieved with this ``oracle'' summarization are better than the LLM summarization by 6\% for both seen and unseen objects. This result suggests that the LLM summarizations are already quite good, and that improvements to LLM summarization could enable further gains for our method in the future.

\begin{table}
\caption{Comparison of different LLMs}
\centering
\small
\setlength\tabcolsep{0.4em}
\begin{tabular}{l|cc|cc}
\toprule
\multirow{2}{*}{Model} & \multicolumn{2}{c|}{Commonsense} & \multicolumn{2}{c}{Summarization} \\
& seen & unseen & seen & unseen \\
\midrule
text-davinci-003 & 45.0\% & 45.6\% & \textbf{91.8\%} & \textbf{91.2\%} \\
text-davinci-002 & 41.8\% & 37.5\% & 84.1\% & 75.7\% \\
code-davinci-002 & 41.4\% & 39.4\% & 88.6\% & 83.2\% \\
PaLM 540B & \textbf{45.5\%} & \textbf{49.6\%} & 84.6\% & 75.7\% \\
\bottomrule
\end{tabular}
\label{tab:llm-comparison}
\end{table}

\mysubsection{Different LLMs.}
Table~\ref{tab:llm-comparison} reports our performance on the benchmark using different LLMs.
We find that \texttt{text-davinci-002} and \texttt{code-davinci-002}~\citep{chen2021evaluating}, which are older variants of GPT-3, are not as good as the newest one (\texttt{text-davinci-003}).
In particular, there is a much larger gap between seen and unseen objects.
This is because the older models are more likely to generate summaries that list out individual objects in the seen set, which does not generalize well to the unseen objects.
For PaLM 540B~\citep{chowdhery2022palm}, we find that while it shows slightly higher performance on commonsense reasoning, it does not do as well as \texttt{text-davinci-003} on summarization, particularly in scenarios where there is a larger number of receptacles to choose from.

\subsection{Human evaluation}

\begin{table*}
\caption{User study results by sorting criteria}
\centering
\small
\setlength\tabcolsep{1em}
\begin{tabular}{l|ccccc|c}
\toprule
Method & Category & Attribute & Function & Subcategory & Multiple & Overall \\
\midrule
CLIP embeddings      & 19.7\%          & 23.7\%          & 11.2\%          & 22.6\%          & 21.2\%          & 19.1\% \\
Summarization (ours) & \textbf{47.4\%} & \textbf{41.9\%} & \textbf{60.0\%} & \textbf{46.1\%} & \textbf{40.6\%} & \textbf{46.9\%} \\
Equally preferred    & 32.9\%          & 34.4\%          & 28.8\%          & 31.3\%          & 38.2\%          & 34.1\% \\
\bottomrule
\end{tabular}
\label{tab:user-study}
\end{table*}

\begin{figure}
\centering
\includegraphics[width=\columnwidth]{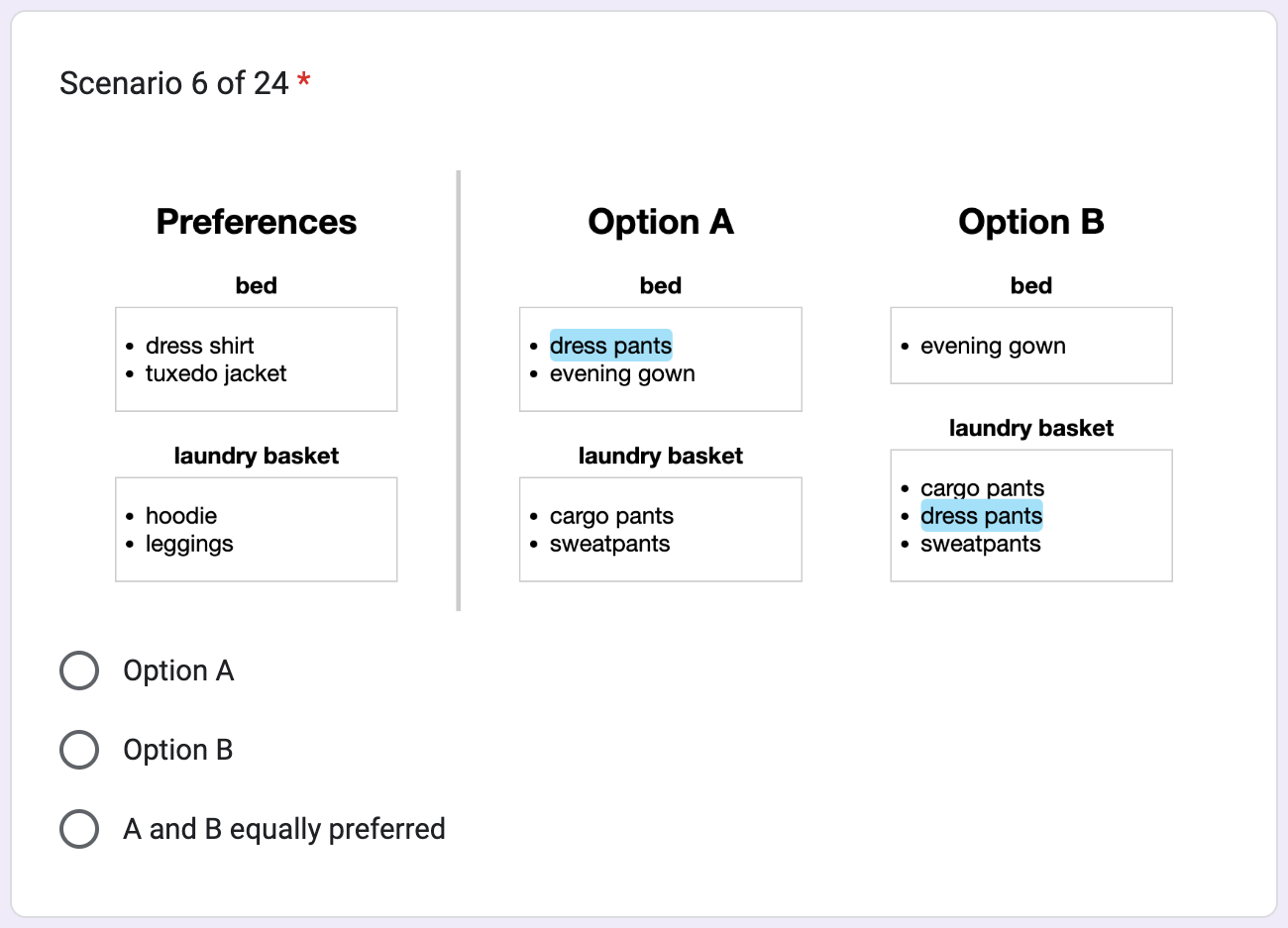}
\caption{\textbf{Example user study question.} This screenshot shows an example survey question from our user study. On the left are preferences, on the right are two placement options corresponding to the two methods being compared. The participant is asked to select the option that is best aligned with the given preferences.}
\label{fig:user-study}
\end{figure}

To evaluate whether humans prefer the preferences learned by our method, we conduct a user study based on the scenarios in our benchmark dataset.
The study asks participants to compare the object placements generated by our method to those of CLIP embeddings, which is the strongest baseline.
The study has 2 objectives:
\begin{enumerate}
\item Evaluate whether humans prefer the object placements generated by our LLM summarization method over those of the CLIP embeddings baseline
\item Evaluate whether human-preferred object placements align with the ground truth placements in our benchmark
\end{enumerate}

\mysubsection{Study setup.}
We recruited 40 participants (24 males and 16 females) consisting of affiliates from author institutions and asked them to fill out an online survey.
Each participant was assigned 24 scenarios randomly selected from the 96 scenarios in the benchmark.
Each scenario in the benchmark is evaluated by 10 participants, giving 960 evaluations in total.

For each scenario, we provide (i) example placements of ``seen'' objects indicating user preferences, and (ii) placements of ``unseen'' objects from both our LLM summarization method and the CLIP embeddings method (example shown in Fig.~\ref{fig:user-study}).
The participant is then asked to specify which of the two object placement options better aligns with the given preferences, or if they are equally preferable.
For the convenience of the participants, we highlight the object placements that differ between the two methods.
We randomize the order of scenarios as well as the order of methods for each scenario (the participant is unaware of which option goes with which method).
For some of the scenarios, both methods give the exact same object placements, so we preselect the third ``equally preferred'' option and exclude them from the surveys given to participants.

\mysubsection{Results.}
Our results across all 960 evaluations are shown in Tab.~\ref{tab:user-study}.
Overall, we find that our LLM summarization method is preferred over the CLIP embeddings baseline 46.9\% of the time, whereas the baseline is preferred 19.1\% of the time, and both methods are equally preferred 34.1\% of the time.
When considering the results broken down by sorting criteria, we find that our method performs particularly well relative to the baseline for the function criteria (\eg formal vs. casual clothes).
Even though the corresponding benchmark accuracy is relatively high (CLIP embeddings in Tab.~\ref{tab:baseline-comparison-criteria}), the baseline method usually sorts by object category (as described in Sec.~\ref{sec:baseline-comparisons}) which can lead to egregiously wrong placements (\eg store dress pants with sweatpants) when the intended sorting criteria is function.

We ran a statistical analysis with the following null hypothesis (H0): There is no significant difference between the preference for our method versus the baseline method.
In other words, the mean fraction of time participants prefer our method over the baseline is equal to 0.5.
For each study participant, we calculated the fraction of time our method was preferred over the baseline method across the 24 scenarios for that participant.
For scenarios where both methods were equally preferred, we gave them both equal weight.
We then conducted a paired t-test, and found a significant difference between our method and the baseline method, with a calculated t-statistic of 9.93 (df = 39), $p < 0.001$, indicating strong evidence to reject the null hypothesis and suggesting that the observed difference in human preference between our method and the baseline is unlikely to have occurred due to random chance.

We also evaluate how well the participant responses align with the ground truth in our benchmark.
For each scenario, we identify which of the two methods is closer to the benchmark ground truth based on unseen object placement accuracy on that scenario.
We then calculate the percent of human responses that prefer the method that is closer to the ground truth.
Overall, across the 40 participants, we find that human responses were aligned with benchmark ground truth $82.2\% \pm 7.7\%$ of the time, or $95.4\% \pm 4.1\%$ if ``equally preferred'' is treated as a wildcard.

\subsection{Real-world experiments}

In our final set of experiments, we test the proposed approach on a robot performing a cleanup task in the real world (Fig.~\ref{fig:teaser}).   
The robot base is a holonomic vehicle capable of any 3-degree-of-freedom motion on the ground plane. This maneuverability comes from the vehicle's Powered-Caster Drive System~\citep{holmberg2000development}, which consists of four caster wheels that are powered to roll and steer as needed to achieve the desired vehicle motion.
The robot manipulator is a Kinova Gen3 \mbox{7-DoF} arm mounted on top of the mobile base with a Robotiq 2F-85 parallel jaw gripper as its end effector.

The robot is placed inside a room with various objects and receptacles on the floor and is then tasked with picking up all the objects and putting them into the correct receptacles according to user preferences. The preferences are provided as a set of textual examples for a particular user (as in the benchmark).
As described in Sec.~\ref{sec:system} and illustrated in Fig.~\ref{fig:overview}, the robot iteratively locates the closest object on the floor, navigates to it, recognizes its category, picks it up, determines the appropriate receptacle for the object, navigates to the receptacle, and then puts the object inside.

\mysubsection{Implementation.}
The robot uses two overhead cameras for 2D robot pose estimation ($x$, $y$, $\theta$) and 2D object localization ($x$, $y$).
The pose of the robot base is estimated using ArUco fiducial markers~\citep{garrido2014automatic} mounted on its top plate (see Fig.~\ref{fig:teaser}). 
The object locations are detected in the overhead camera using ViLD~\citep{gu2021open}, while the receptacle locations are hard-coded for each scenario.
We found that these design choices work well for our mobile robot system.
However, other pose trackers and object detectors could also be used instead.

To navigate in the scene, the robot calculates the shortest collision-free path to the target position using an occupancy map that includes obstacles in the scene such as receptacles.
It then uses the pure pursuit algorithm~\citep{coulter1992implementation} to follow the computed path.

After the robot arrives at the closest object, it uses a camera mounted on its base (and pointed forward at the ground) to take a close-up, centered image of the object, then determines the object category using cosine similarity between text and image features in the CLIP embedding space~\citep{radford2021learning}.
The set of object categories in the LLM summary is automatically extracted and used as the target label set for CLIP.
Note that without these categories from LLM summarization, a human would have to manually specify a list of fine-grained object classes potentially present in the target scene in order to use CLIP for object classification.

After the object category is identified, the system uses the LLM summarization to predict the appropriate receptacle and manipulation primitive for the object.
The robot then moves the object into the receptacle with a sequence of two high-level manipulation primitives: (i) pick and (ii) place or toss.
The ``pick'' primitive uses the gripper to grasp at the center of the detected object.
The ``place'' primitive moves the gripper to a location just above the selected receptacle and drops the grasped object in.
The ``toss'' primitive swings the robot arm and releases the gripper with timing that results in tossing~\citep{zeng2020tossingbot} of the grasped object into the selected receptacle. 

\begin{figure*}
\centering
\setlength\tabcolsep{1.5pt}
\begin{tabular}{cccc}
\includegraphics[width=0.248\textwidth]{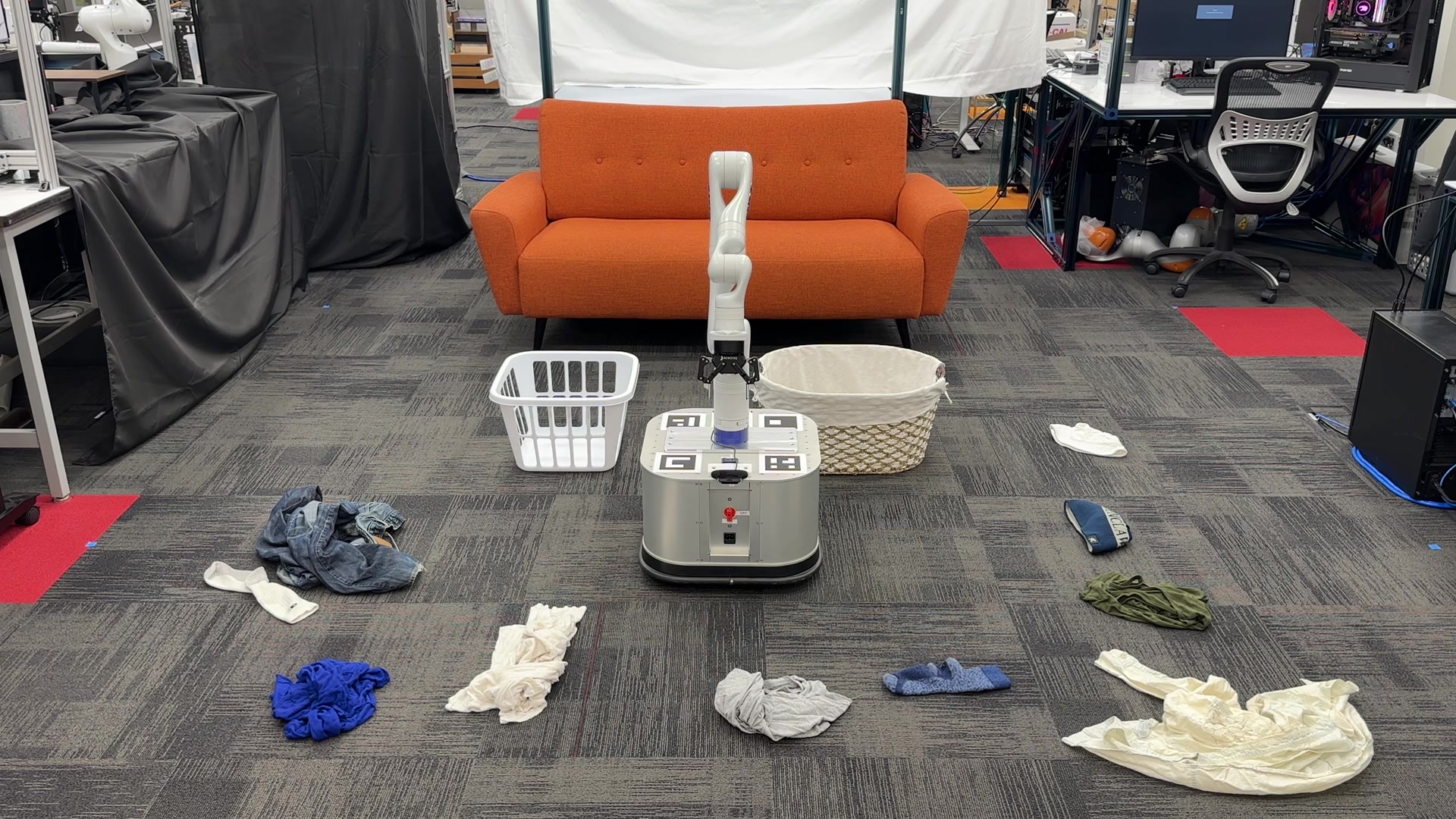} &
\includegraphics[width=0.248\textwidth]{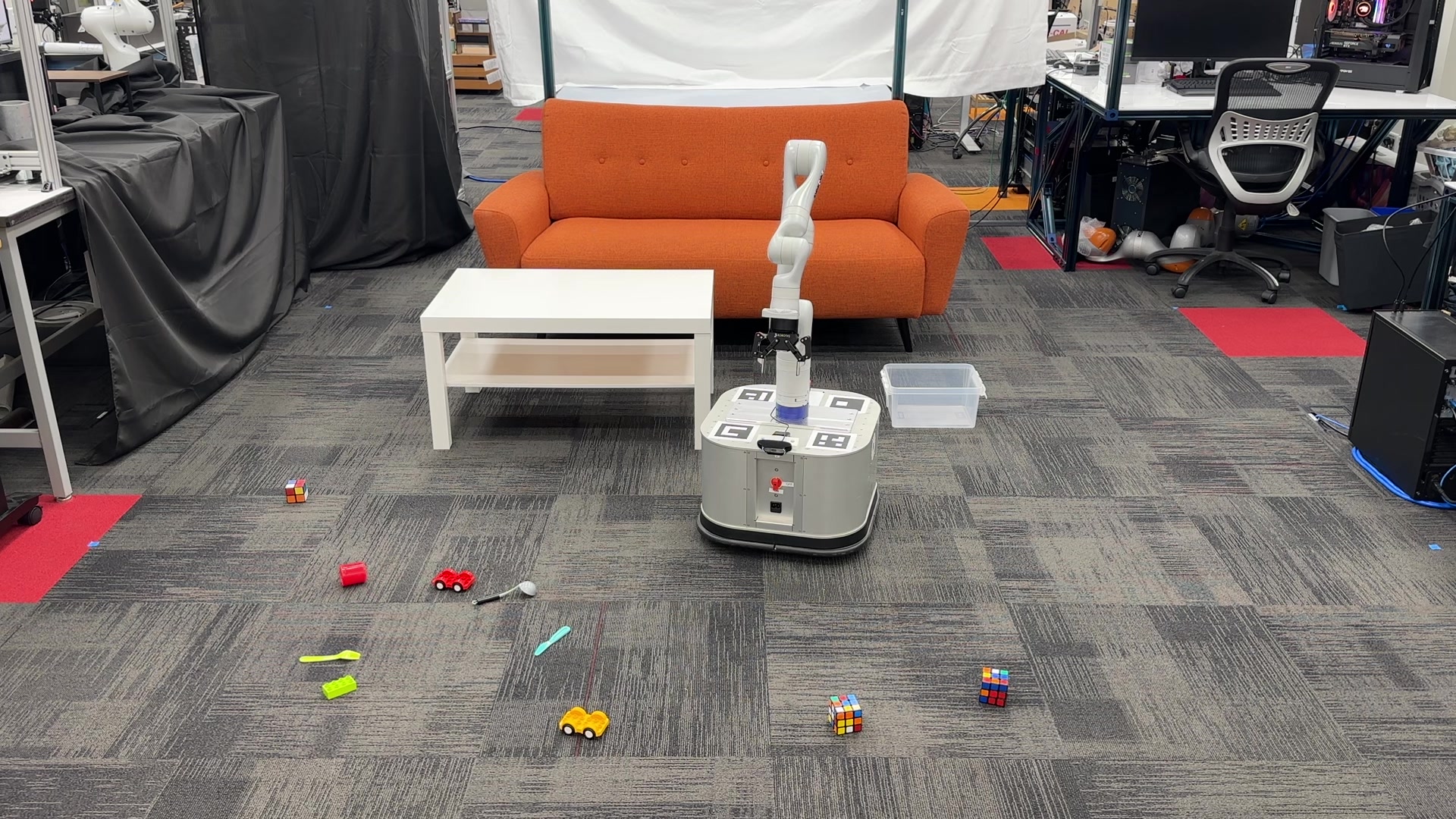} &
\includegraphics[width=0.248\textwidth]{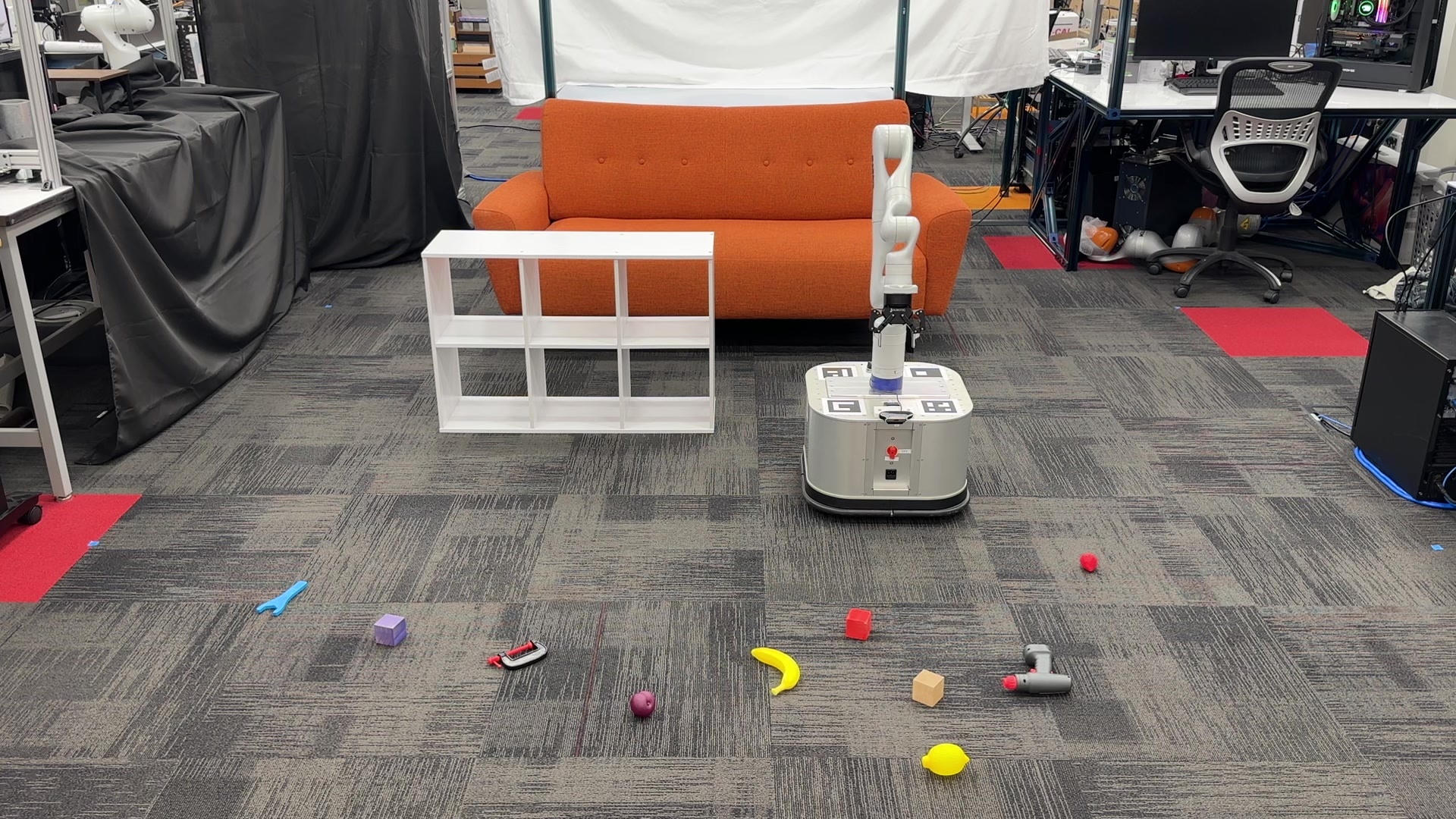} &
\includegraphics[width=0.248\textwidth]{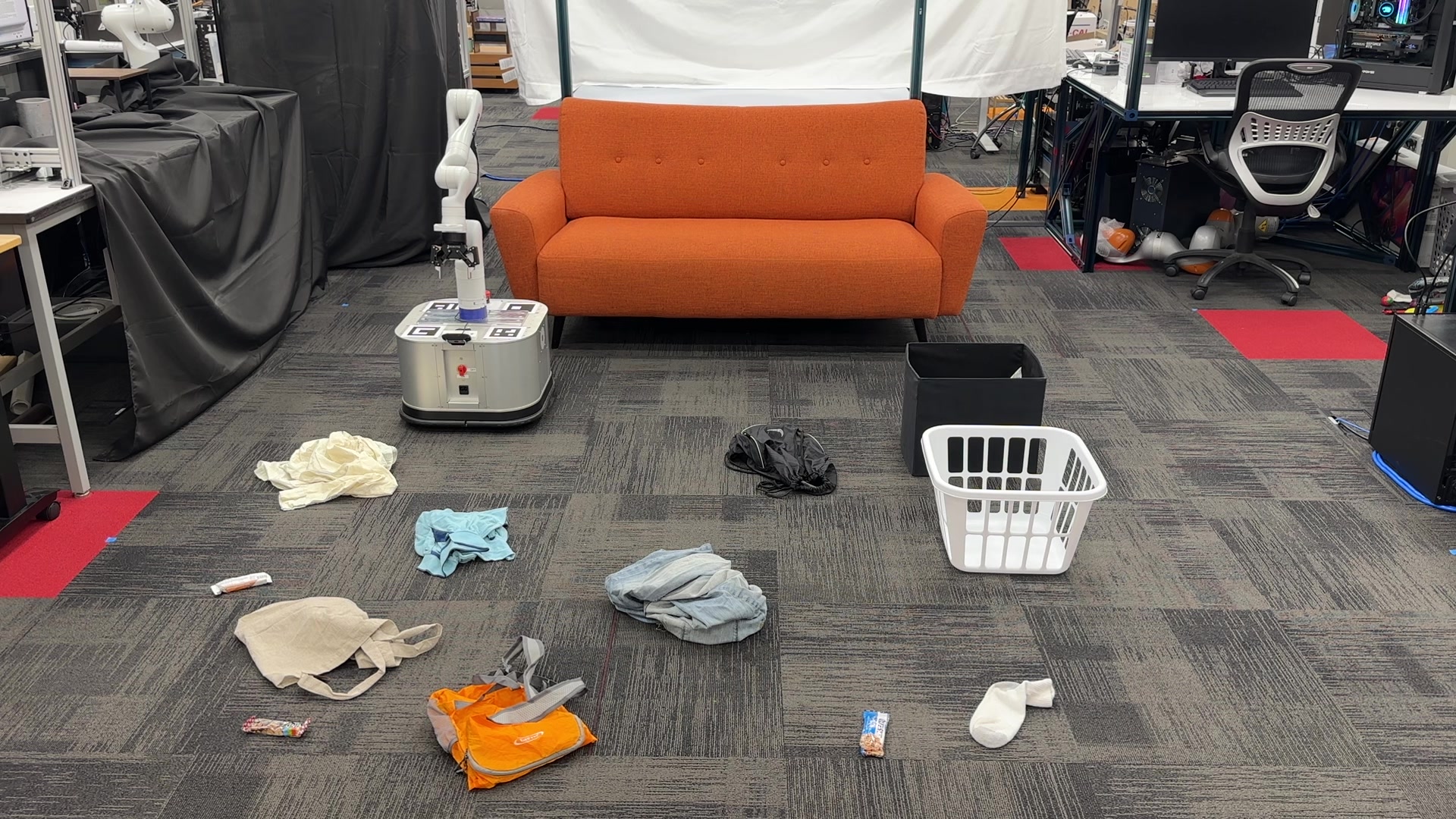} \\
\includegraphics[width=0.248\textwidth]{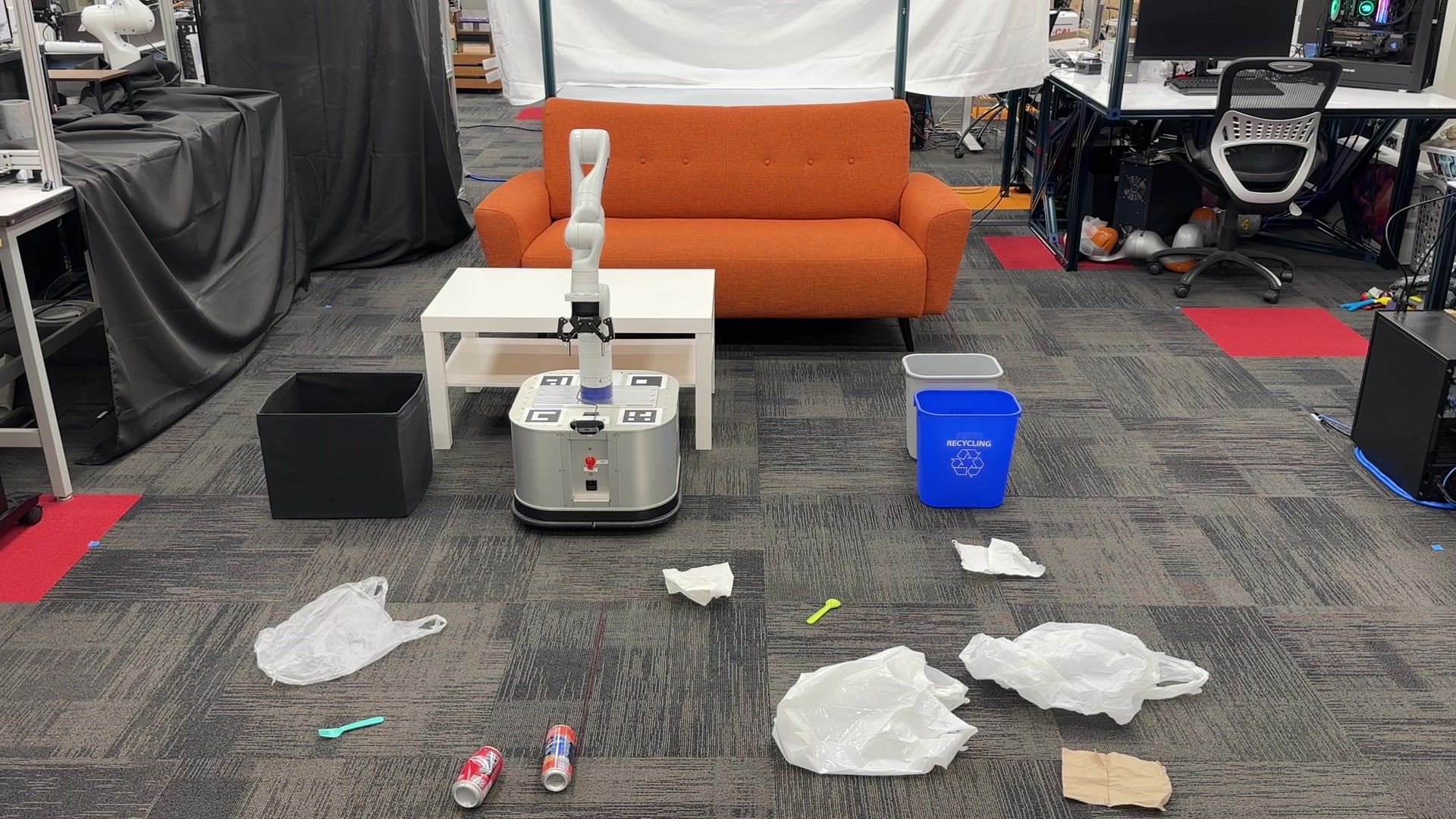} &
\includegraphics[width=0.248\textwidth]{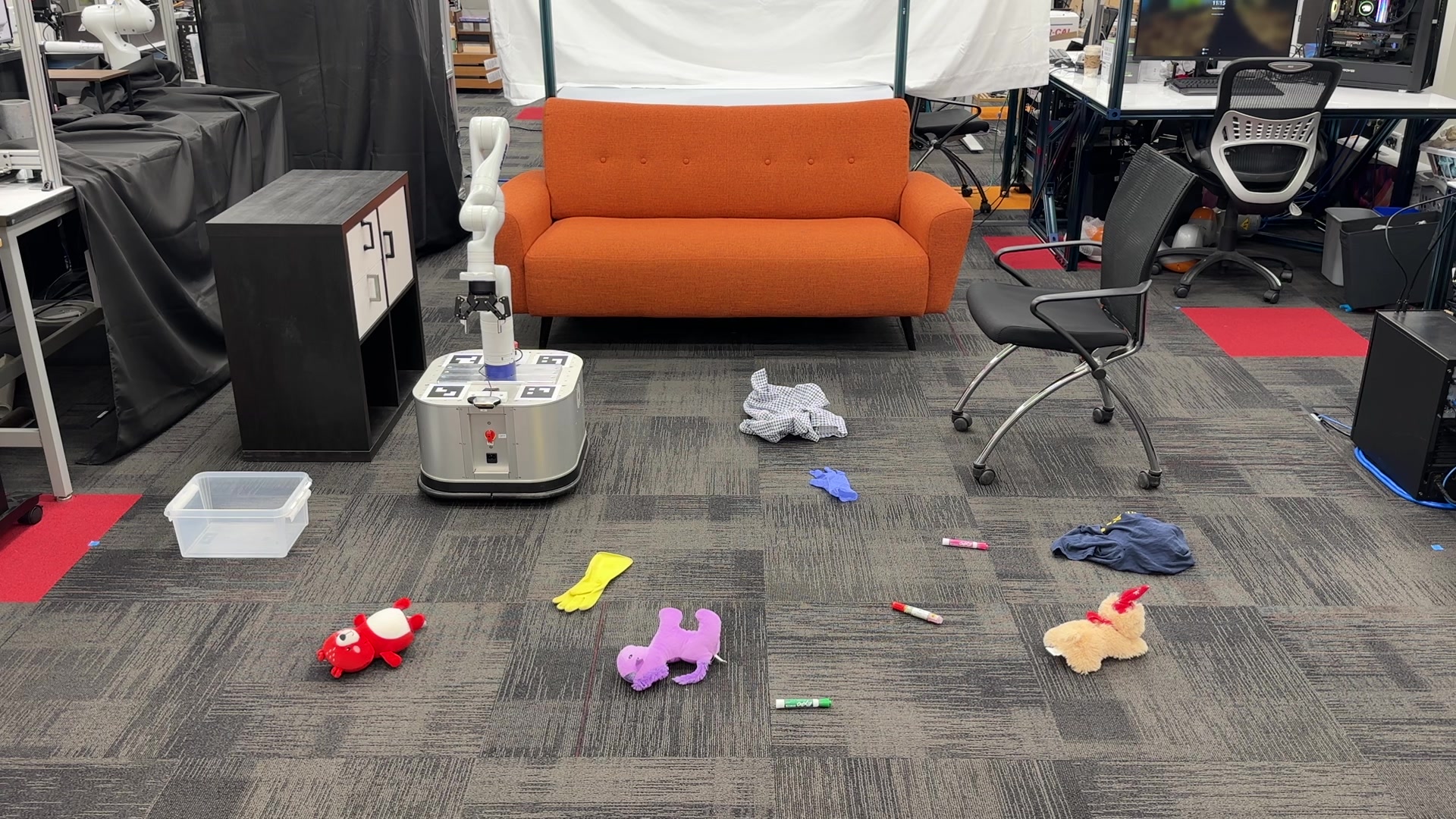} &
\includegraphics[width=0.248\textwidth]{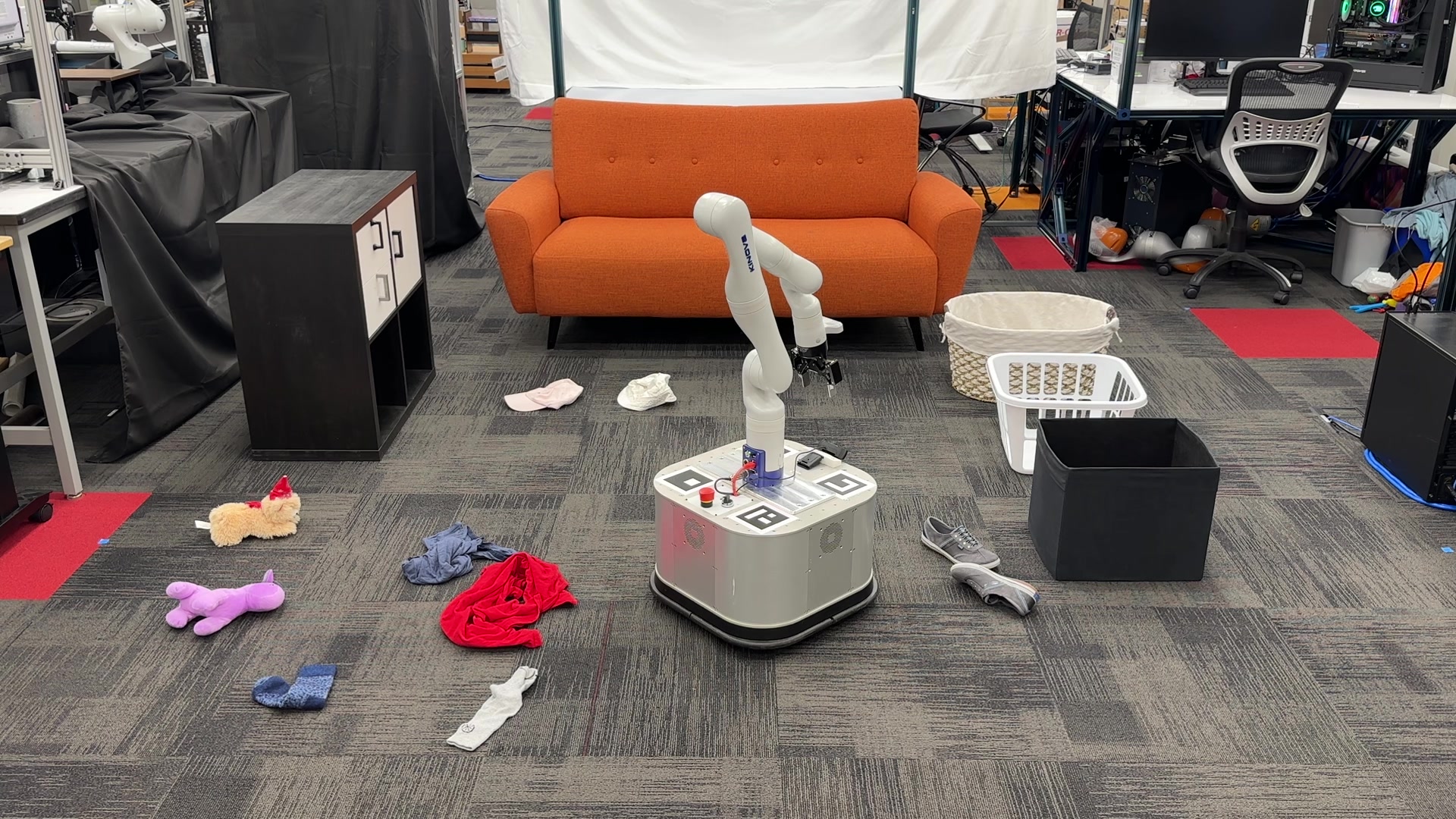} &
\includegraphics[width=0.248\textwidth]{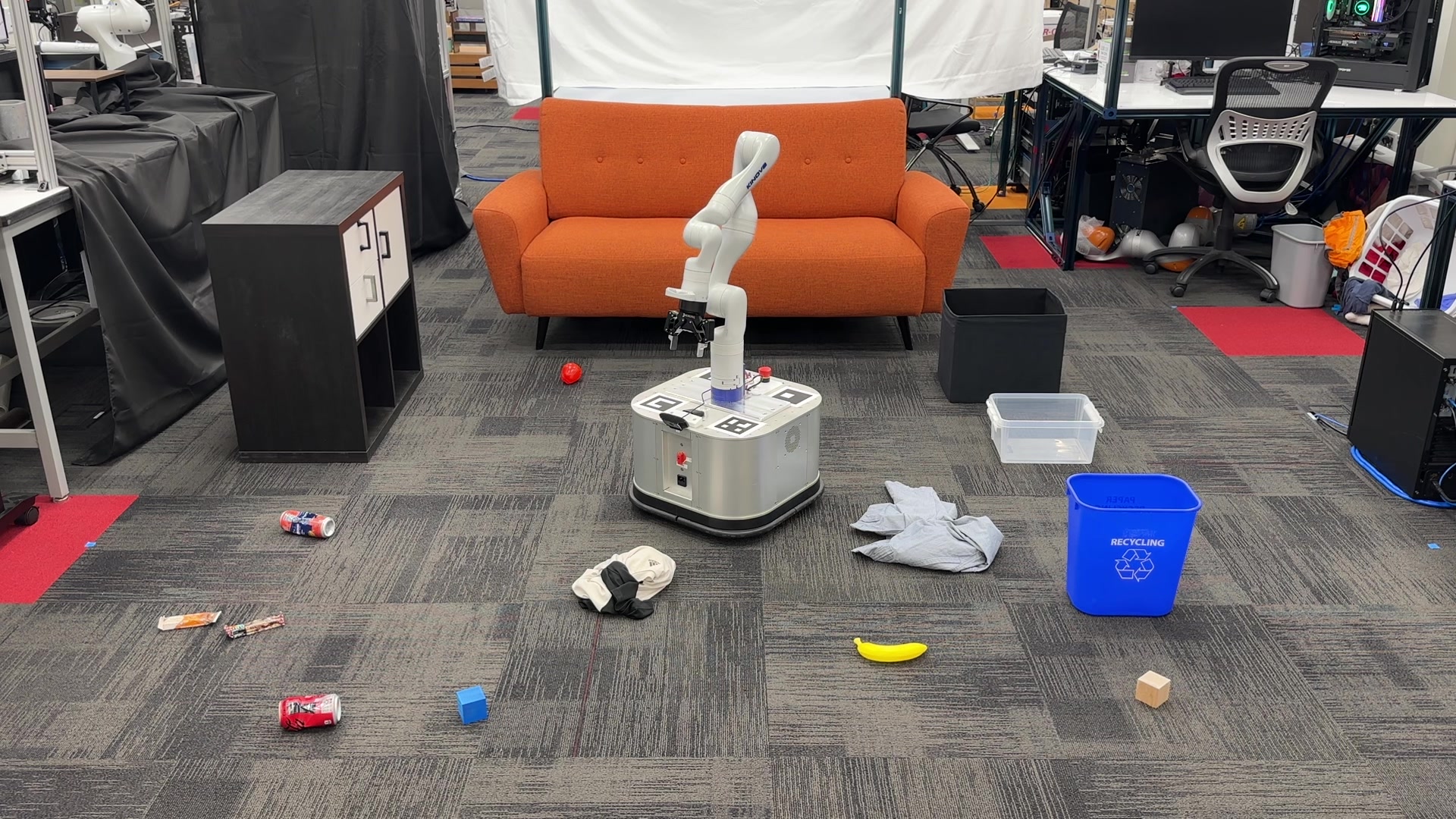} \\
\end{tabular}
\caption{\textbf{Real-world scenarios.} We evaluate our mobile manipulation system in 8 real-world scenarios, encompassing a wide variety of objects and receptacles.}
\label{fig:real-scenarios}
\end{figure*}

\begin{figure}
\centering
\setlength\tabcolsep{1.5pt}
\begin{tabular}{cc}
\includegraphics[width=0.495\columnwidth]{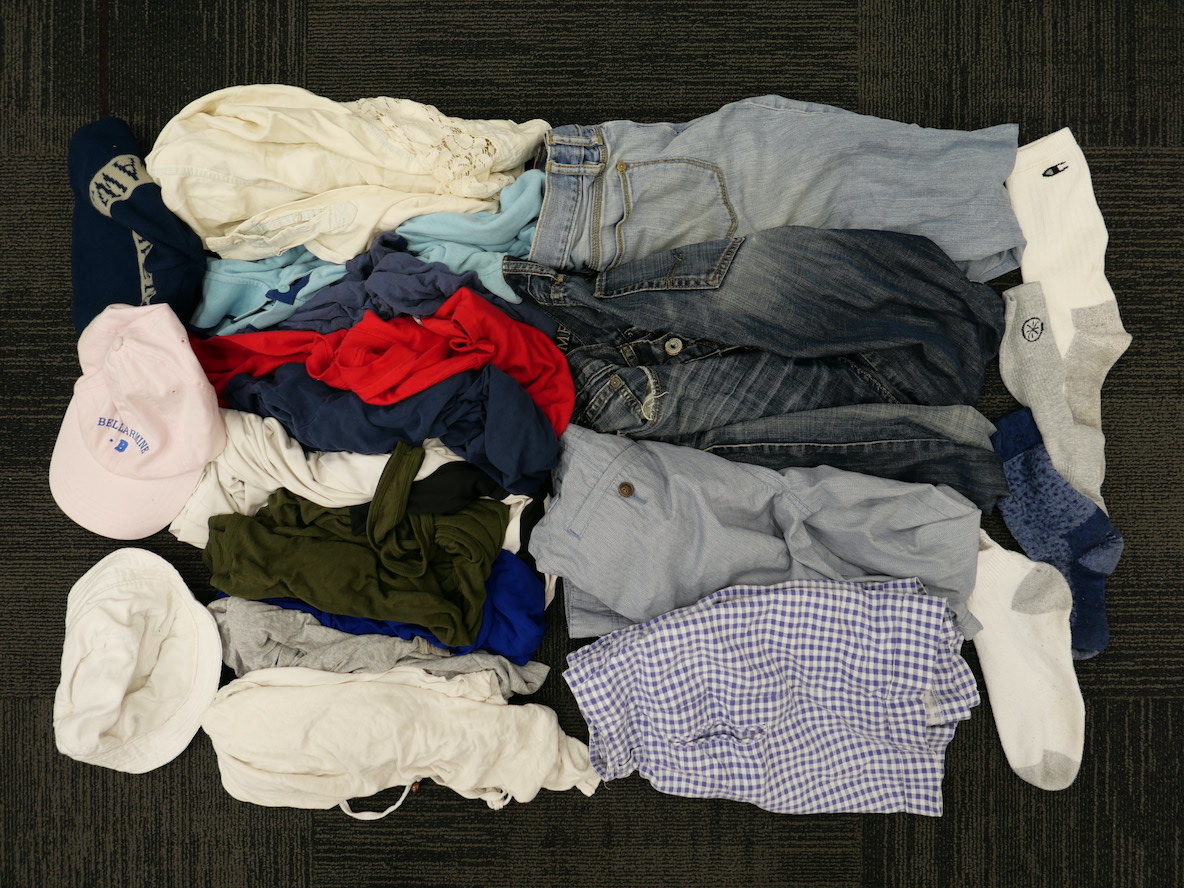} &
\includegraphics[width=0.495\columnwidth]{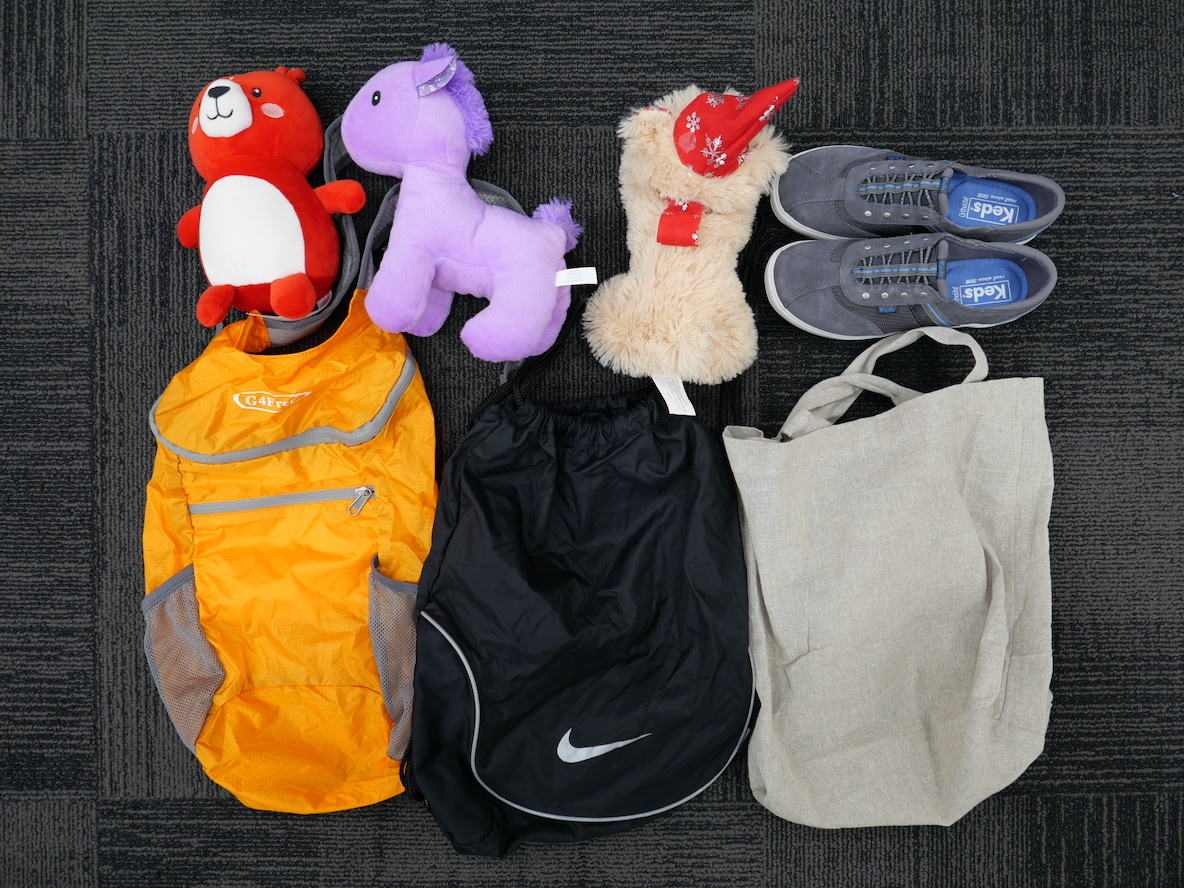} \\
\includegraphics[width=0.495\columnwidth]{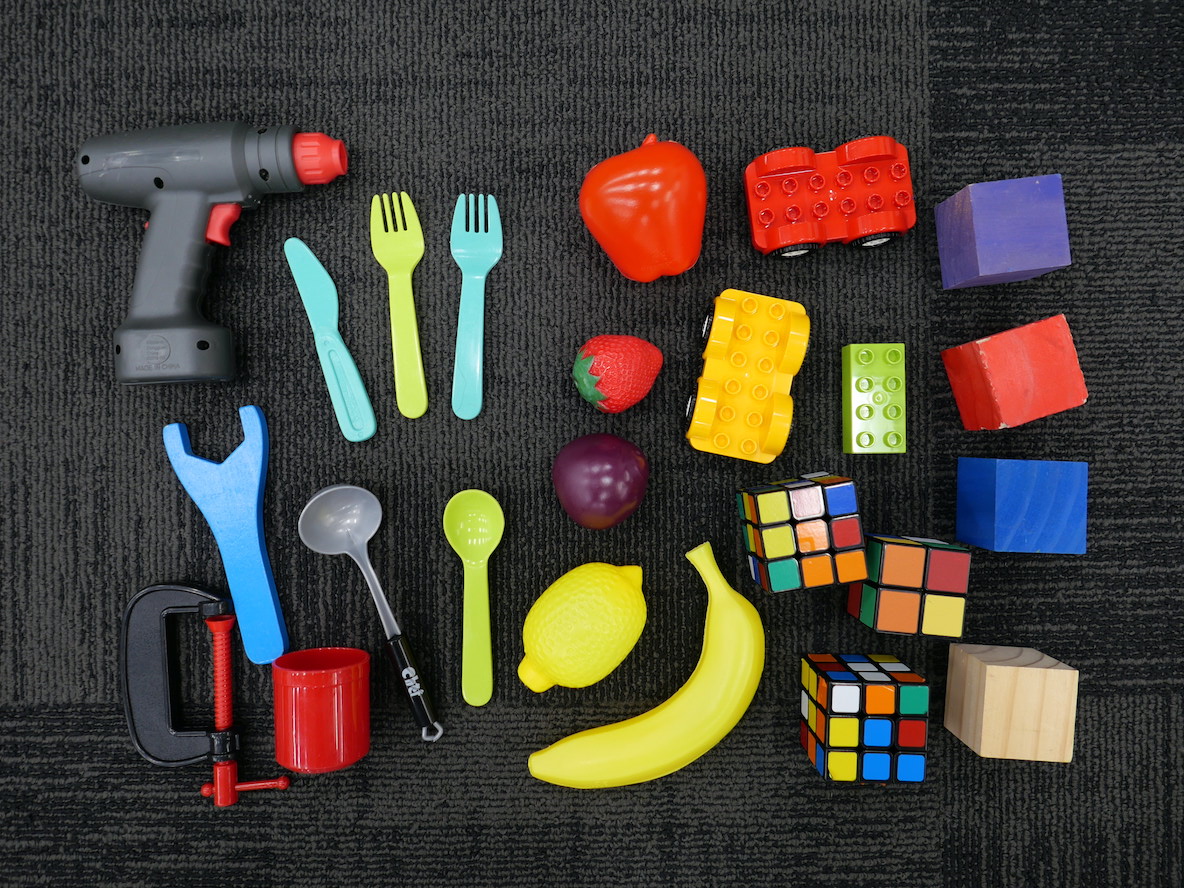} &
\includegraphics[width=0.495\columnwidth]{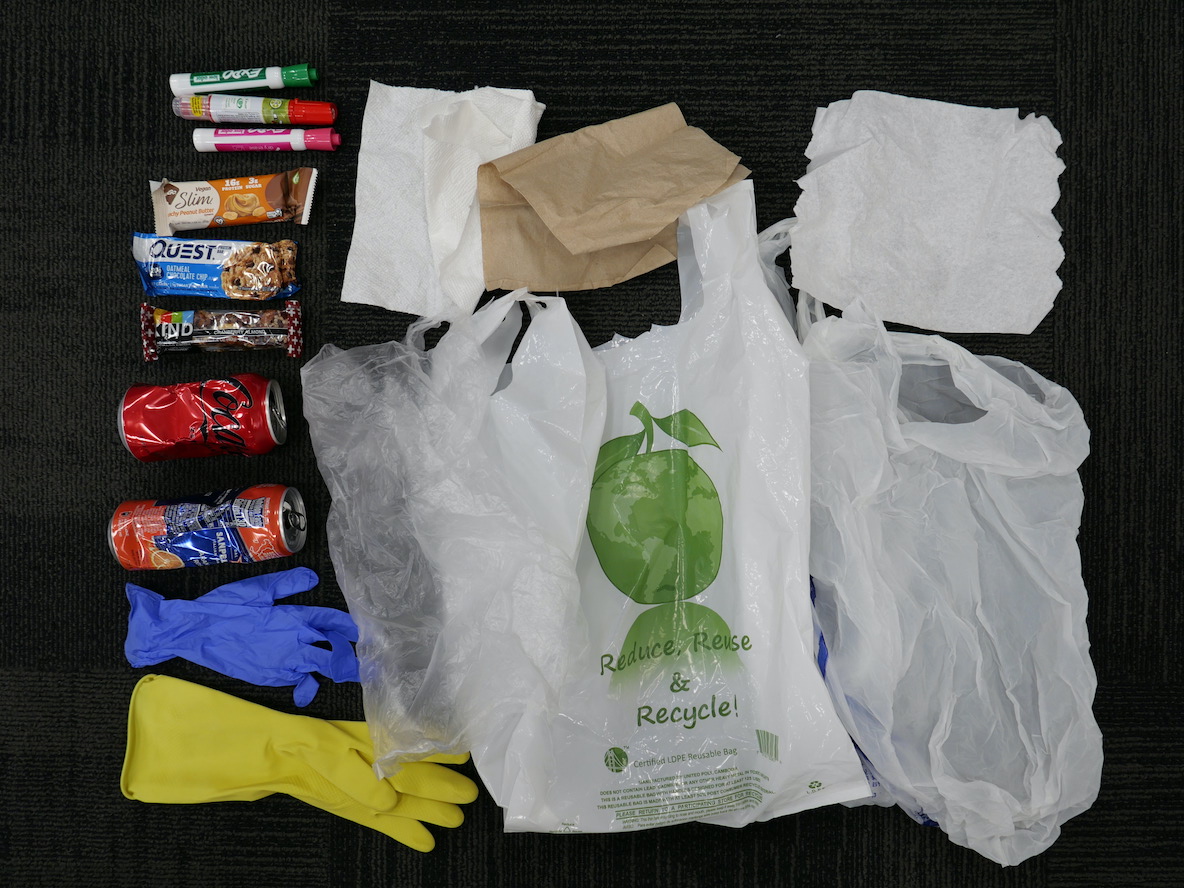} \\
\end{tabular}
\caption{\textbf{Real-world objects.} 70 unique ``unseen'' test objects are represented in our real-world scenarios.}
\label{fig:real-objects}
\end{figure}

\begin{figure}
\centering
\setlength\tabcolsep{1.5pt}
\begin{tabular}{cc}
\includegraphics[width=0.495\columnwidth]{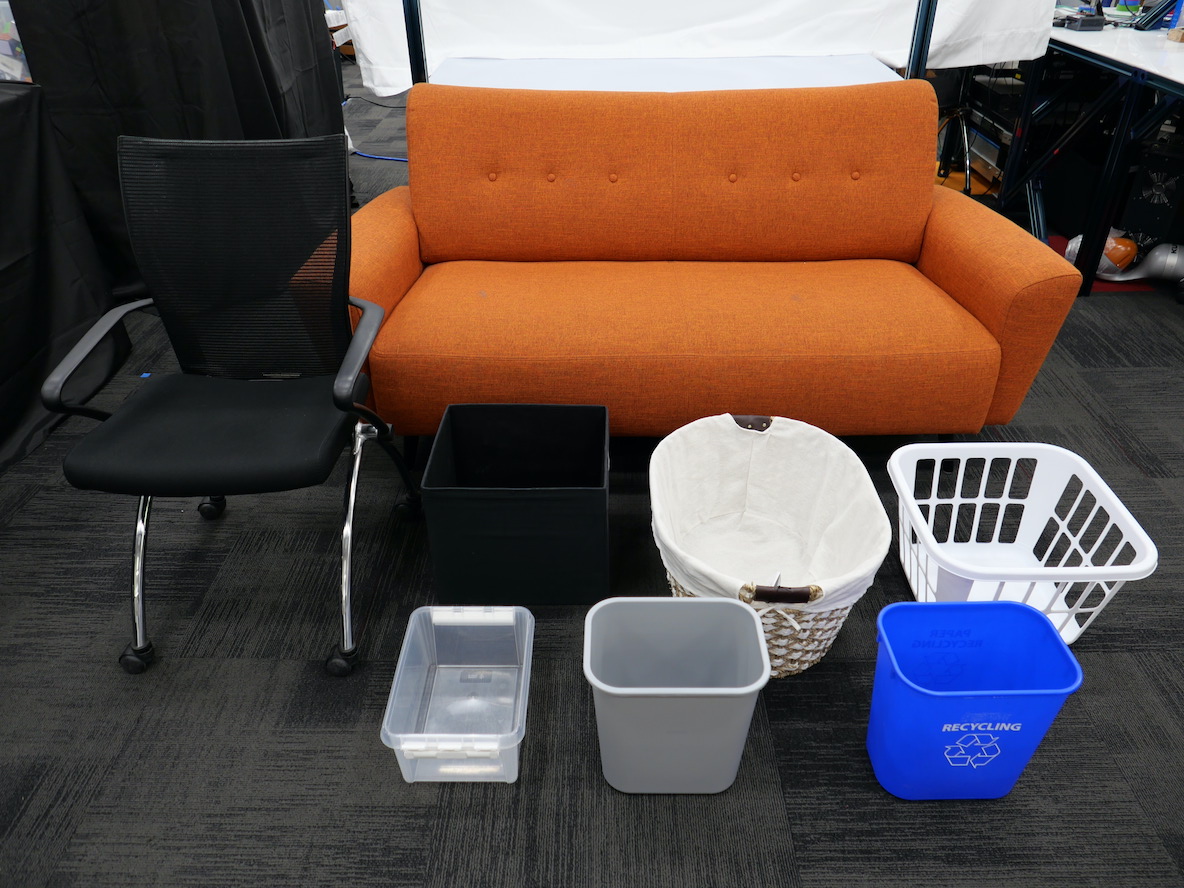} &
\includegraphics[width=0.495\columnwidth]{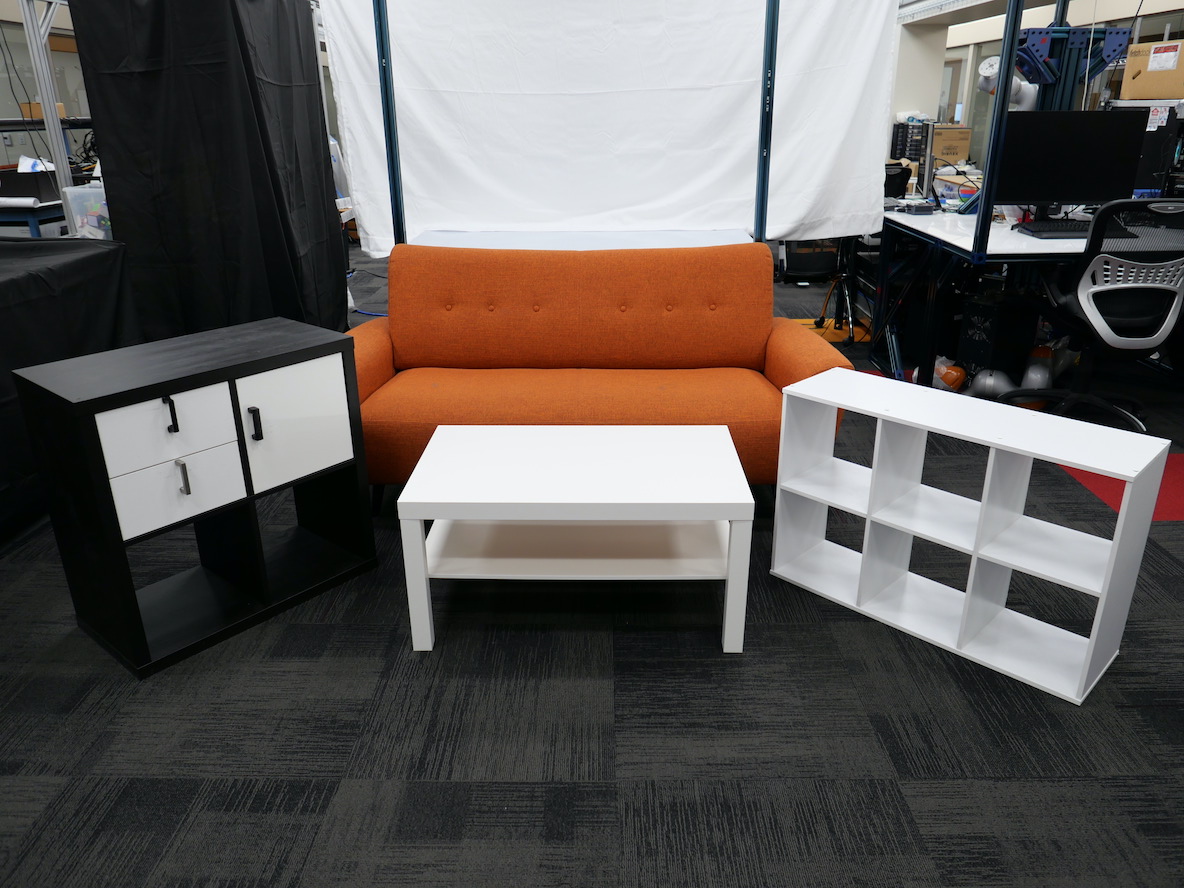} \\
\end{tabular}
\caption{\textbf{Real-world receptacles.} 11 unique receptacles are represented in our real-world scenarios.}
\label{fig:real-receptacles}
\end{figure}

\mysubsection{Real-world evaluation.}
Using this mobile robot system, we ran tests on 8 real-world scenarios as shown in Fig.~\ref{fig:real-scenarios}, each with its own set of 10 objects, 2--5 receptacles, 4--10 ``seen'' examples indicating preferences for which objects should go into which receptacles and which primitive should be used to put them there, as well as 10 ``unseen'' test objects. Across all 8 scenarios, 70 unique ``unseen'' test objects (Fig.~\ref{fig:real-objects}) and 11 unique receptacles (Fig.~\ref{fig:real-receptacles}) are represented.

For each scenario, we asked the robot to perform 3 runs of the cleanup task and measured its success throughout operation.
Overall, the system was able to put 85.0\% of the objects into the correct receptacle during these tests. For qualitative examples, please refer to the supplementary material and additional videos at \mbox{\url{https://tidybot.cs.princeton.edu}}.

Looking at the results in more detail, there were 240 objects to be cleaned up in total (8 scenarios, 10 objects per scenario, 3 runs per scenario).
We observed that the overhead camera was able to localize 92.5\% of the objects, and the object classifier correctly identified the object category for 95.5\% of the localized objects.
Given the predicted object category, the LLM selected the appropriate receptacle and manipulation primitive for 100\% of localized objects.
Additionally, the robot succeeded in executing the chosen primitive for 96.2\% of the localized objects.
In terms of speed, the robot took on average 15--20 seconds to pick up and put away each object.

\mysubsection{Visual language model (VLM) evaluation.}
In this section, we perform a quantitative comparison of different visual language models (VLMs) within our real robot system.
Recall that for each object successfully localized by the overhead camera, the real robot will first use its egocentric camera to take a close-up image of the object before picking it up.
This image is given to a VLM to determine the category of the object.
To conduct our analysis, we save all egocentric images from our real world evaluation (222 in total across all test runs) and annotate them.

To evaluate a VLM, we run all 222 images through the model and determine the fraction of images in which the centered foreground object is correctly recognized.
We compare along two axes: (i) model type and (ii) vocabulary used for the target label set.
The model types we consider (all open-vocabulary) are (i) CLIP~\citep{radford2021learning}, which was the image classifier used in our final system, and two alternatives, (ii) ViLD~\citep{gu2021open} and (iii) OWL-ViT~\citep{minderer2022simple}.
The vocabulary options we consider are (i) the set of categories output by LLM summarization (\eg clothing, fruit, ...), which was used in our final system, (ii) a list of human-annotated names for all objects in the current scenario (\eg blue jeans, apple, ...), and (iii) a list of human-annotated object names across all scenarios (instead of just one scenario).
Note that the human-annotated options for the vocabulary are for analysis only, as it would be infeasible to ask a human to annotate every object encountered during robot operation.
Results are shown in Tab.~\ref{tab:vlm-comparison}.

\begin{table}
\caption{Comparison of different VLMs}
\centering
\small
\setlength\tabcolsep{0.4em}
\begin{tabular}{l|ccc}
\toprule
& CLIP & ViLD & OWL-ViT \\
\midrule
Summarized categories & \textbf{95.5\%} & 76.1\% & 45.9\% \\
Scenario object names & \textbf{70.7\%} & 59.9\% & 24.8\% \\
All object names      & \textbf{52.3\%} & 36.5\% & 18.5\% \\
\bottomrule
\end{tabular}
\label{tab:vlm-comparison}
\end{table}

Looking at the results comparing different VLMs (columns of Tab.~\ref{tab:vlm-comparison}), we find that CLIP performs the best out of all the models.
One reason is that CLIP will always output a prediction, whereas the object detectors (ViLD and OWL-ViT) will sometimes detect no objects in the image.
Additionally, ViLD and OWL-ViT are derived from CLIP, and it is possible that the process of adapting the models to localize bounding boxes degrades their performance on object classification.

Qualitatively, the main failure mode of CLIP is reporting the class of an object in the background rather than that of the foreground object.
This is expected since CLIP performs an image-wide classification.
We also observe that CLIP is often not able to consider noun phrases as complete units.
For example, the phrase ``white socks'' may match strongly with anything that looks white.

For ViLD and OWL-ViT (both object detectors), we use the bounding box closest to the center of the image as the detection, since the egocentric camera is pointed directly at the pick location on the floor.
We expected that this localization would improve accuracy since foreground objects can be isolated from background objects (unlike with CLIP).
However, we find that quantitatively, both ViLD and OWL-ViT perform worse than CLIP.
Qualitatively, ViLD works well with small rigid objects, but struggles with larger deformable objects (such as clothes or stuffed animals), outputting many extraneous detections corresponding to parts of objects.
Additionally, for both ViLD and OWL-ViT, we find that the foreground object is sometimes not detected at all, even though it is always prominently placed in the center of the image.

When looking at results for different vocabularies (rows of Tab.~\ref{tab:vlm-comparison}), we find that using the categories from the LLM summary performs the best.
This is partly because the VLM has to differentiate between a much smaller number of options (2--5 categories vs. 10 or 65 object names).
Note again that the use of object names is not actually feasible in a real system due to the human annotation burden.
By contrast, our use of LLM summarized categories allows the system to directly generalize to novel objects as the VLM only needs to correctly identify the closest category rather than what the specific object is.

\subsection{Limitations}

\mysubsection{LLM summarization.}
While LLMs are generally able to summarize preferences well, we find that there are still cases in which the generated summary is not quite right.
The most common failure mode is when the generated summary simply lists out the seen objects rather than summarizing into categories.
Summaries of that nature are too specific and do not generalize well to unseen objects.
Another failure mode is when the LLM summarizes receptacles by grouping them together (\eg top drawer and bottom drawer might be summarized as drawers), resulting in poor performance when using the summary for receptacle selection.

\mysubsection{Real-world system.}
Our implementation of the real-world system contains simplifications such as the use of hand-written manipulation primitives, use of top-down grasps, and assumption of known receptacle locations.
These limitations could be addressed by incorporating more advanced primitives into our system and expanding the capabilities of the perception system.
Additionally, since the mobile robots cannot drive over objects, the system would not work well in excessive clutter.
It would be interesting to incorporate more advanced high-level planning, so that instead of always picking up the closest object, the robot could reason about whether it needs to first clear itself a path to move through the clutter.

\section{Conclusion} 

In this work, we showed that the summarization capabilities of large language models (LLMs) can be used to generalize user preferences for personalized robotics.
Given a handful of example preferences for a particular person, we use LLM summarization to infer a generalized set of rules to manipulate objects according to the user's preferences.
We show that our summarization approach outperforms several strong baselines on our benchmark, and we also evaluate our approach on a real-world mobile manipulator called TidyBot, which can successfully clean up test scenarios with a success rate of 85.0\%.
Our approach provides a promising direction for developing personalized robotic systems that can learn generalized user preferences quickly and effectively from only a small set of examples.
Unlike classical approaches that require costly data collection and model training, we show that LLMs can be directly used off-the-shelf to achieve generalization in robotics, leveraging the powerful summarization capabilities they have learned from vast amounts of text data.

\backmatter

\section*{Acknowledgments}

The authors would like to thank William Chong, Kevin Lin, and Jingyun Yang for fruitful technical discussions, and Bob Holmberg for mentorship and support in building up the mobile platforms. This work was supported in part by the Princeton School of Engineering, Toyota Research Institute, and the National Science Foundation under CCF-2030859, DGE-1656466, and IIS-2132519.

\begin{appendices}
\section{LLM prompts}
\label{sec:appendix-full-prompts}

This section contains the full prompts used for all LLM text completion tasks.
Each prompt consists of 1--3 in-context examples in \textcolor{light-gray}{gray} followed by a test example that we ask the LLM to complete.
The portion of the test example that is generated by the LLM is \colorbox{highlight}{highlighted}.
We use the same in-context examples across all scenarios in both the benchmark and the real-world system.
For each scenario, only the final test example is modified.

\subsection{Summarization for receptacle selection}

\codeblock{\textcolor{light-gray}{objects = ["dried figs", "protein bar", "cornmeal", "Macadamia nuts", "vinegar", "herbal tea", "peanut oil", "chocolate bar", "bread crumbs", "Folgers instant coffee"]\\
receptacles = ["top rack", "middle rack", "table", "shelf", "plastic box"]\\
pick\_and\_place("dried figs", "plastic box")\\
pick\_and\_place("protein bar", "shelf")\\
pick\_and\_place("cornmeal", "top rack")\\
pick\_and\_place("Macadamia nuts", "plastic box")\\
pick\_and\_place("vinegar", "middle rack")\\
pick\_and\_place("herbal tea", "table")\\
pick\_and\_place("peanut oil", "middle rack")\\
pick\_and\_place("chocolate bar", "shelf")\\
pick\_and\_place("bread crumbs", "top rack")\\
pick\_and\_place("Folgers instant coffee", "table")\\
\# Summary: Put dry ingredients on the top rack, liquid ingredients in the middle rack, tea and coffee on the table, packaged snacks on the shelf, and dried fruits and nuts in the plastic box.\\
\strut\\
objects = ["yoga pants", "wool sweater", "black jeans", "Nike shorts"]\\
receptacles = ["hamper", "bed"]\\
pick\_and\_place("yoga pants", "hamper")\\
pick\_and\_place("wool sweater", "bed")\\
pick\_and\_place("black jeans", "bed")\\
pick\_and\_place("Nike shorts", "hamper")\\
\# Summary: Put athletic clothes in the hamper and other clothes on the bed.\\
\strut\\
objects = ["Nike sweatpants", "sweater", "cargo shorts", "iPhone", "dictionary", "tablet", "Under Armour t-shirt", "physics homework"]\\
receptacles = ["backpack", "closet", "desk", "nightstand"]\\
pick\_and\_place("Nike sweatpants", "backpack")\\
pick\_and\_place("sweater", "closet")\\
pick\_and\_place("cargo shorts", "closet")\\
pick\_and\_place("iPhone", "nightstand")\\
pick\_and\_place("dictionary", "desk")\\
pick\_and\_place("tablet", "nightstand")\\
pick\_and\_place("Under Armour t-shirt", "backpack")\\
pick\_and\_place("physics homework", "desk")\\
\# Summary: Put workout clothes in the backpack, other clothes in the closet, books and homeworks on the desk, and electronics on the nightstand.
\strut\\}}

\codeblock{objects = ["jacket", "candy bar", "soda can", "Pepsi can", "jeans", "wooden block", "orange", "chips", "wooden block 2", "apple"]\\
receptacles = ["recycling bin", "plastic storage box", "black storage box", "sofa", "drawer"]\\
pick\_and\_place("jacket", "sofa")\\
pick\_and\_place("candy bar", "plastic storage box")\\
pick\_and\_place("soda can", "recycling bin")\\
pick\_and\_place("Pepsi can", "recycling bin")\\
pick\_and\_place("jeans", "sofa")\\
pick\_and\_place("wooden block", "drawer")\\
pick\_and\_place("orange", "black storage box")\\
pick\_and\_place("chips", "plastic storage box")\\
pick\_and\_place("wooden block 2", "drawer")\\
pick\_and\_place("apple", "black storage box")\\
\# Summary:\llmcompletion{ Put clothes on the sofa, snacks in }\\
\llmcompletion{the plastic storage box, cans in the recycling }\\
\llmcompletion{bin, wooden blocks in the drawer, and fruits }\\
\llmcompletion{in the black storage box.}}

\subsection{Receptacle selection}

\codeblock{\textcolor{light-gray}{\# Summary: Put clothes in the laundry basket and toys in the storage box.\\
objects = ["socks", "toy car", "shirt", "Lego brick"]\\
receptacles = ["laundry basket", "storage box"]\\
pick\_and\_place("socks", "laundry basket")\\
pick\_and\_place("toy car", "storage box")\\
pick\_and\_place("shirt", "laundry basket")\\
pick\_and\_place("Lego brick", "storage box")}\\
\strut\\
\# Summary: Put clothes on the sofa, snacks in the plastic storage box, cans in the recycling bin, wooden blocks in the drawer, and fruits in the black storage box.\\
objects = ["jacket", "candy bar", "soda can", "Pepsi can", "jeans", "wooden block", "orange", "chips", "wooden block 2", "apple"]\\
receptacles = ["recycling bin", "plastic storage box", "black storage box", "sofa", "drawer"]\\
pick\_and\_place("jacket",\llmcompletion{ "sofa")}\\
\llmcompletion{pick\_and\_place("candy bar", "plastic storage }\\
\llmcompletion{box")}\\
\llmcompletion{pick\_and\_place("soda can", "recycling bin")}\\
\llmcompletion{pick\_and\_place("Pepsi can", "recycling bin")}\\
\llmcompletion{pick\_and\_place("jeans", "sofa")}\\
\llmcompletion{pick\_and\_place("wooden block", "drawer")}\\
\llmcompletion{pick\_and\_place("orange", "black storage box")}\\
\llmcompletion{pick\_and\_place("chips", "plastic storage box")}\\
\llmcompletion{pick\_and\_place("wooden block 2", "drawer")}\\
\llmcompletion{pick\_and\_place("apple", "black storage box")}}

\subsection{Summarization for primitive selection}

\codeblock{\textcolor{light-gray}{objects = ["granola bar", "hat", "toy car", "Lego brick", "fruit snacks", "shirt"]\\
pick\_and\_toss("granola bar")\\
pick\_and\_place("hat")\\
pick\_and\_place("toy car")\\
pick\_and\_place("Lego brick")\\
pick\_and\_toss("fruit snacks")\\
pick\_and\_place("shirt")\\
\# Summary: Pick and place clothes and toys, pick and toss snacks.}\\
\strut\\
objects = ["jacket", "candy bar", "soda can", "Pepsi can", "jeans", "wooden block", "orange", "chips", "wooden block 2", "apple"]\\
pick\_and\_place("jacket")\\
pick\_and\_toss("candy bar")\\
pick\_and\_toss("soda can")\\
pick\_and\_toss("Pepsi can")\\
pick\_and\_place("jeans")\\
pick\_and\_place("wooden block")\\
pick\_and\_toss("orange")\\
pick\_and\_toss("chips")\\
pick\_and\_place("wooden block 2")\\
pick\_and\_toss("apple")\\
\# Summary:\llmcompletion{ Pick and place clothes and wooden }\\
\llmcompletion{blocks, pick and toss snacks and drinks.}}

\subsection{Primitive selection}

\codeblock{\textcolor{light-gray}{\# Summary: Pick and place clothes, pick and toss snacks.\\
objects = ["granola bar", "hat", "toy car", "Lego brick", "fruit snacks", "shirt"]\\
pick\_and\_toss("granola bar")\\
pick\_and\_place("hat")\\
pick\_and\_place("toy car")\\
pick\_and\_place("Lego brick")\\
pick\_and\_toss("fruit snacks")\\
pick\_and\_place("shirt")\\
\strut\\
\# Summary: Pick and place granola bars, hats, toy cars, and Lego bricks, pick and toss fruit snacks and shirts.\\
objects = ["clothing", "snack"]\\
pick\_and\_place("clothing")\\
pick\_and\_toss("snack")}\\
\strut\\
\# Summary: Pick and place clothes and wooden blocks, pick and toss snacks and drinks.\\
objects = ["jacket", "candy bar", "soda can", "Pepsi can", "jeans", "wooden block", "orange", "chips", "wooden block 2", "apple"]\\
\llmcompletion{pick\_and\_place("jacket")}\\
\llmcompletion{pick\_and\_place("jeans")}\\
\llmcompletion{pick\_and\_place("wooden block")}\\
\llmcompletion{pick\_and\_place("wooden block 2")}\\
\llmcompletion{pick\_and\_toss("candy bar")}\\
\llmcompletion{pick\_and\_toss("soda can")}\\
\llmcompletion{pick\_and\_toss("Pepsi can")}\\
\llmcompletion{pick\_and\_toss("orange")}\\
\llmcompletion{pick\_and\_toss("chips")}\\
\llmcompletion{pick\_and\_toss("apple")}}

\subsection{Category extraction for real-world system}

\codeblock{\textcolor{light-gray}{\# Summary: Put shirts on the bed, jackets and pants on the chair, and bags on the shelf.\\
objects = ["shirt", "jacket or pants", "bag"]\\
\strut\\
\# Summary: Put pillows on the sofa, clothes on the chair, and shoes on the rack.\\
objects = ["pillow", "clothing", "shoe"]}\\
\strut\\
\# Summary: Put clothes on the sofa, snacks in the plastic storage box, cans in the recycling bin, wooden blocks in the drawer, and fruits in the black storage box.\\
objects = ["\llmcompletion{clothing", "snack", "can", }\\
\llmcompletion{"wooden block", "fruit"]}}

\subsection{Receptacle selection for real-world system}

\codeblock{\textcolor{light-gray}{\# Summary: Put clothes in the laundry basket and toys in the storage box.\\
objects = ["socks", "toy car", "shirt", "Lego brick"]\\
receptacles = ["laundry basket", "storage box"]\\
pick\_and\_place("socks", "laundry basket")\\
pick\_and\_place("toy car", "storage box")\\
pick\_and\_place("shirt", "laundry basket")\\
pick\_and\_place("Lego brick", "storage box")}\\
\strut\\
\# Summary: Put clothes on the sofa, snacks in the plastic storage box, cans in the recycling bin, wooden blocks in the drawer, and fruits in the black storage box.\\
objects = ["clothing", "snack", "can", "wooden block", "fruit"]\\
receptacles = ["recycling bin", "plastic storage box", "black storage box", "sofa", "drawer"]\\
pick\_and\_place("clothing",\llmcompletion{  "sofa")}\\
\llmcompletion{pick\_and\_place("snack", "plastic storage box")}\\
\llmcompletion{pick\_and\_place("can", "recycling bin")}\\
\llmcompletion{pick\_and\_place("wooden block", "drawer")}\\
\llmcompletion{pick\_and\_place("fruit", "black storage box")}}

\subsection{Primitive selection for real-world system}

\codeblock{\textcolor{light-gray}{\# Summary: Pick and place clothes, pick and toss snacks.\\
objects = ["granola bar", "hat", "toy car", "Lego brick", "fruit snacks", "shirt"]\\
pick\_and\_toss("granola bar")\\
pick\_and\_place("hat")\\
pick\_and\_place("toy car")\\
pick\_and\_place("Lego brick")\\
pick\_and\_toss("fruit snacks")\\
pick\_and\_place("shirt")\\
\strut\\
\# Summary: Pick and place granola bars, hats, toy cars, and Lego bricks, pick and toss fruit snacks and shirts.\\
objects = ["clothing", "snack"]\\
pick\_and\_place("clothing")\\
pick\_and\_toss("snack")}\\
\strut\\
\# Summary: Pick and place clothes and wooden blocks, pick and toss snacks and drinks.\\
objects = ["clothing", "snack", "can", "wooden block", "fruit"]\\
\llmcompletion{pick\_and\_place("clothing")}\\
\llmcompletion{pick\_and\_place("wooden block")}\\
\llmcompletion{pick\_and\_toss("snack")}\\
\llmcompletion{pick\_and\_toss("can")}\\
\llmcompletion{pick\_and\_toss("fruit")}}

\end{appendices}

\bibliography{references}

\end{document}